\documentclass[preprint,12pt,authoryear]{elsarticle}

\usepackage{amsmath,amssymb}
\usepackage{booktabs}
\usepackage{graphicx}
\usepackage{subcaption}
\usepackage{multirow}
\usepackage{array}
\usepackage{threeparttable}
\usepackage{xcolor}
\usepackage{hyperref}
\usepackage{lineno}

\journal{ISPRS Journal of Photogrammetry and Remote Sensing}

\biboptions{authoryear,round}

\begin{document}

\begin{frontmatter}

\title{Multi-level latent-diffusion adaptation for cross-region point cloud segmentation of ancient terraces}

\author[liacs]{Sina Mohammadi\corref{cor1}}
\ead{s.mohammadi@liacs.leidenuniv.nl}
\author[delft]{Roderik Lindenbergh}
\author[upf]{Javier Ruiz-Pérez}
\author[upf,icrea,gaes]{Marco Madella}
\author[liacs,delft]{Sebastian Fajardo}
\cortext[cor1]{Corresponding author.}
\affiliation[liacs]{organization={Leiden Institute of Advanced Computer Science (LIACS), Leiden University}, addressline={Einsteinweg 55, 2333 CA}, city={Leiden}, country={The Netherlands}}
\affiliation[delft]{organization={Geoscience \& Remote Sensing Department, Delft University of Technology}, addressline={Stevinweg 1, 2628 CN Delft}, city={Delft}, country={The Netherlands}}
\affiliation[upf]{organization={Department of Humanities, Universitat Pompeu Fabra},addresline={carrer de la Mercé, 08002},city={Barcelona},country={Spain}}
\affiliation[icrea]{organization={ICREA},addresline={Pg. Lluís Companys 23, 08010},city={Barcelona},country={Spain}}
\affiliation[gaes]{organization={School of Geography, Archaeology, and Environmental Studies, University of the Witwatersrand},addresline={1 Jan Smuts Avenue, Braamfontein 2000},city={Johannesburg},country={South Africa}}

\begin{abstract}
Automated lidar feature detection can support archaeological surveys, but models trained in one landscape may transfer poorly to another because of differences in terrain, vegetation, acquisition characteristics, and feature morphology. We address this problem for cross-region detection of ancient terraces using transductive unsupervised domain adaptation that exploits an unlabelled target survey. We propose MELD (Multi-level Embedding-space Latent Diffusion), which trains separate diffusion denoisers on target-domain representations at four hierarchical PointNet++ feature levels. These target distribution priors are then used to refine source representations during supervised training and target representations during inference, allowing adaptation to act directly on the multi-scale features consumed by the segmentation decoder. We additionally investigate local geometric descriptors derived from K-nearest-neighbour neighbourhoods, including surface normals, slope, and aspect. MELD is evaluated bidirectionally between the Teyuna and Miraflores lidar surveys in the Sierra Nevada de Santa Marta, Colombia, against a source-only model and nine configurations of established domain adaptation approaches. MELD raises F1 over the source-only model from 0.397 to 0.459 for Teyuna-to-Miraflores transfer and from 0.557 to 0.594 in the reverse direction, achieving the highest F1 among the evaluated adaptation methods in both directions. Adding local geometry further raises F1 to 0.542 and 0.606, respectively. Multi-level ablation shows that refining all four PointNet++ levels provides the strongest performance in both directions. These results show that explicit modelling of target feature distributions in hierarchical point cloud feature space, complemented by local geometry, provides an effective route to cross-region archaeological mapping without target labels.

\end{abstract}


\end{frontmatter}


\section{Introduction}
\label{sec:introduction}

Airborne and uncrewed aerial system (UAS) lidar have substantially expanded the possibilities for archaeological prospection by enabling detailed three-dimensional mapping of terrain and anthropogenic landforms across large and densely vegetated landscapes \citep{vinci2025,stular2021airborne,mazzacca2022seeing,freeland2016automated}. Ancient agricultural and residential terraces are particularly important archaeological features because their distribution can provide evidence of settlement organisation, cultivation practices, slope modification, and water management. Their identification, however, remains challenging. Terrace risers may be eroded or partially buried, vegetation and post-depositional processes may obscure their morphology, and their size, orientation, relief, and topographic context can vary considerably between landscapes. As lidar coverage continues to expand, exhaustive manual interpretation therefore becomes an increasingly important bottleneck.

Deep learning has become an important approach for the semantic analysis of airborne laser scanning (ALS) point clouds, with recent methods increasingly learning discriminative representations directly from irregular three-dimensional point sets. For example, density-aware convolutional architectures have been developed to account for spatially varying point density while capturing local and contextual information for ALS point cloud classification \citep{li2020dance}, while active and incremental learning strategies have been investigated to reduce the substantial annotation effort required for supervised semantic segmentation of large-scale ALS datasets \citep{lin2020active}. These developments demonstrate both the capacity of deep networks to learn informative representations directly from point clouds and the continuing challenge of obtaining sufficient labelled training data. In archaeology and cultural heritage, deep learning has shown considerable potential for feature detection from lidar-derived raster representations, including digital terrain models and local-relief visualisations \citep{verschoofvandervaart2019,verschoof2020combining,trier2019using,https://doi.org/10.1002/arp.1806}. Related studies targeting archaeological water-management features and agricultural terraces have likewise relied primarily on lidar-derived variables and orthophotos \citep{tiwari2023waterharvesting,tubog2025terraces}. In parallel, point-based deep-learning approaches have been applied directly to three-dimensional cultural-heritage point clouds for semantic segmentation \citep{pierdicca2020point,haznedar2023implementing,yang2023heritagepointcloud}. Among architectures designed to operate directly on point sets, PointNet and PointNet++ are particularly attractive for terrace detection because they operate on unordered 3D point sets: PointNet learns permutation-invariant point-set representations, while PointNet++ extends this paradigm with hierarchical aggregation of local neighbourhoods \citep{qi2017pointnet,qi2017pointnetplusplus}. This direct point-based representation also avoids imposing an explicit rasterisation step before learning. Their learned representations can nonetheless become domain-specific. A detector trained in one survey may need to be applied to another where terrain, vegetation, acquisition configuration, point density, point cloud statistics, archaeological prevalence, and feature morphology differ. Such changes alter the input and feature distributions and can cause a model that performs well in the source region to transfer poorly to a geographically distinct target landscape.

This domain shift limits the scalability of supervised deep-learning approaches for archaeological mapping because applying them to new surveys may require new expert-labelled training data. Producing such annotations requires substantial expert effort and is time-consuming. In contrast, the complete lidar survey of a new region is often available before detailed archaeological annotation begins. This creates a practically important setting in which labelled data are available from a previously mapped source domain, while large quantities of unlabelled data are available from the new target domain. The central problem is therefore not only how to construct an accurate terrace detector, but how to exploit the unlabelled target survey itself to transfer that detector across regions without requiring target annotations.

Unsupervised domain adaptation (UDA) addresses this problem by adapting a model learned from a labelled source domain to the distribution of an unlabelled target domain without requiring target annotations \citep{wang2018survey}. Among widely used UDA strategies, source-target representation alignment and target-dependent regularisation provide two important mechanisms for exploiting unlabelled target data. Statistical alignment approaches reduce differences between source and target feature distributions using criteria such as maximum mean discrepancy or correlation alignment \citep{long2015dan,sun2016coral,sun2016deepcoral}, while adversarial approaches encourage the network to learn representations from which source and target samples cannot readily be distinguished \citep{ganin2016dann,long2018cdan, liu2021adversarial}. In point cloud analysis, prior work has jointly aligned local and global point cloud features across multiple scales by combining adaptive local feature alignment with adversarial global alignment \citep{qin2019pointdan}. These approaches can be effective, but they also have limitations. Reducing an aggregate source-target discrepancy may not fully capture the structure of heterogeneous target feature distributions and does not necessarily preserve or improve the task-relevant structure of the feature space. In particular, when domain shift is class dependent, feature distributions are multimodal, or class proportions differ substantially between domains, improved marginal alignment does not guarantee that semantically corresponding source and target representations become better aligned.

This issue is especially relevant to archaeological terrace segmentation. Background terrain strongly dominates the point distribution, while terrace points form a comparatively small and heterogeneous class. A global alignment objective may therefore be influenced primarily by the dominant background distribution. Moreover, the target domain may contain terrace morphologies or terrain configurations that are weakly represented in the source domain. Under such conditions, simply reducing a global discrepancy between the two domains may not fully capture the structure of the target representations that is relevant to segmentation.

Target-dependent regularisation provides another approach to unsupervised domain adaptation. Methods based on entropy minimisation \citep{grandvalet2004,vu2019advent}, consistency or self-ensembling \citep{french2018selfensembling}, and pseudo-label self-training \citep{zou2018cbst, wang2025test} exploit unlabelled target predictions rather than explicitly aligning source and target representations. These approaches can encourage more confident and stable target predictions, but their effectiveness depends in part on the quality of the model's current predictions. When the source-trained model is already biased under domain shift, prediction-based regularisation can reinforce incorrect decisions rather than correct them. This is again potentially important for highly imbalanced segmentation, where uncertain target terrace points may initially be assigned to the dominant background class and subsequently become more confidently misclassified.

The limitations of only reducing a discrepancy between source and target representations and of regularising target predictions, motivate a different question: can the structure of the target feature distribution itself be explicitly learned from the unlabelled target data and then used to refine the representations on which prediction depends?

Diffusion models provide a natural mechanism for addressing this question \citep{sohldickstein2015,ho2020ddpm}, as they avoid relying exclusively on a global discrepancy measure that may not capture the internal structure of the target feature distribution. Denoising diffusion models learn complex data distributions by progressively corrupting samples with noise and training a reverse denoising process to recover samples consistent with the underlying distribution \citep{sohldickstein2015,ho2020ddpm}. In feature space, this denoising process can capture the local structure of complex and potentially multimodal target distributions. In our framework, we exploit this property by learning diffusion priors directly from unlabelled target features and using them to refine the representations supplied to the segmentation network. 

Prior diffusion-based approaches for handling domain shift have operated at the input level, on representations obtained from pretrained generative diffusion models, or at an individual learned feature stage \citep{gao2023dda,shim2024cloudfixer,ji2025diff,peng2024dad,prabhudesai2023diffusion}. Consequently, they do not explicitly address domain shift across the multiple hierarchical feature levels produced by the segmentation backbone. To our knowledge, no prior work has introduced level-specific target latent-diffusion priors to refine multiple hierarchical representations. This distinction is particularly relevant to PointNet++, the backbone adopted in this study, which progressively aggregates local neighbourhoods through a sequence of set-abstraction stages, producing representations at increasingly coarse spatial scales. Shallower representations retain relatively fine information about local geometry, while deeper representations encode increasingly broad and semantic context. For terrace segmentation, both types of information are important: local surface orientation, breaks in slope, and neighbourhood structure contribute to terrace morphology, whereas the spacing, alignment, and grouping of neighbouring terrace elements, together with their relationship to the surrounding terrain, provide broader spatial context. Domain shift can affect these hierarchy levels differently. Changes in point density, local surface roughness, vegetation remnants, and acquisition characteristics may primarily alter relatively shallow representations, whereas differences in terrace morphology, the spatial arrangement of neighbouring terraces, and their surrounding terrain may become more pronounced in deeper features. Applying diffusion-based adaptation at only one feature level may therefore leave domain-specific discrepancies at other levels of the feature hierarchy unaddressed.

To address this gap, we propose learning separate target domain diffusion priors for multiple hierarchical task representations and leveraging them jointly for dense prediction. Specifically, we introduce MELD (Multi-level Embedding-space Latent Diffusion), a transductive framework for point cloud domain adaptation. Given an unlabelled target survey, the PointNet++ encoder generates latent representations at four set-abstraction levels. Instead of learning a single target feature distribution, we train a separate diffusion denoiser on the target domain latent tokens at each level. Each denoiser therefore models the target representation at a different spatial and semantic scale.

During supervised source training, corresponding source features are partially corrupted and refined through short reverse-diffusion trajectories using the target-trained denoisers. The resulting target-guided representations are supplied to their respective locations in the PointNet++ feature-propagation decoder. A straight-through estimator allows the forward pass to use the diffusion-refined representations while retaining segmentation gradients to the original encoder features. During target inference, the same level-specific denoisers refine the hierarchical target features before decoding. MELD therefore adapts the actual representations consumed by the segmentation decoder rather than relying solely on an auxiliary source-target discrepancy or a prediction-based regularisation term.

In addition to learned hierarchical representations, terrace morphology is also expressed through explicit local geometric properties. Surface normals, slope, and aspect can provide complementary information about terrace form and local terrain structure. We therefore also investigate KNN-derived geometric descriptors as auxiliary inputs and assess whether they provide additional gains beyond diffusion-based adaptation.

The contributions of this paper are as follows:

\begin{enumerate}

\item We introduce MELD for unsupervised domain adaptation through level-specific target distribution modelling: separate denoising diffusion models learn target domain feature distributions at multiple hierarchical levels of a segmentation encoder with different spatial and semantic characteristics and use the resulting level-specific priors to refine the corresponding representations before decoding.

\item We show empirically that the multi-level construction, and not merely the use of diffusion, is what produces the gain: Refining all four PointNet++ levels achieves the strongest result in each direction, whereas applying conventional alignment losses at more levels yields no such trend.

\item We combine target-aware latent refinement with explicit local terrain geometry, including KNN-derived surface normals, slope, and sine/cosine aspect, and show that these descriptors provide complementary gains beyond diffusion adaptation alone.

\item We provide a rigorous bidirectional cross-region evaluation against established UDA approaches under a shared PointNet++ backbone. MELD achieves the highest F1 among the evaluated adaptation methods in both transfer directions, while combining MELD with explicit geometric descriptors produces the strongest overall results.

\end{enumerate}

The remainder of the paper describes the study areas and data (Section~\ref{sec:data}) and MELD (Section~\ref{sec:method}), details the experimental protocol (Section~\ref{sec:experiments}), reports the results (Section~\ref{sec:results}), and discusses the findings, limitations, and future research directions (Sections~\ref{sec:discussion}--\ref{sec:limitation}).

\section{Study areas and data}
\label{sec:data}

\subsection{Study regions}
\begin{figure}[t]
    \centering
    \includegraphics[width=0.93\columnwidth]{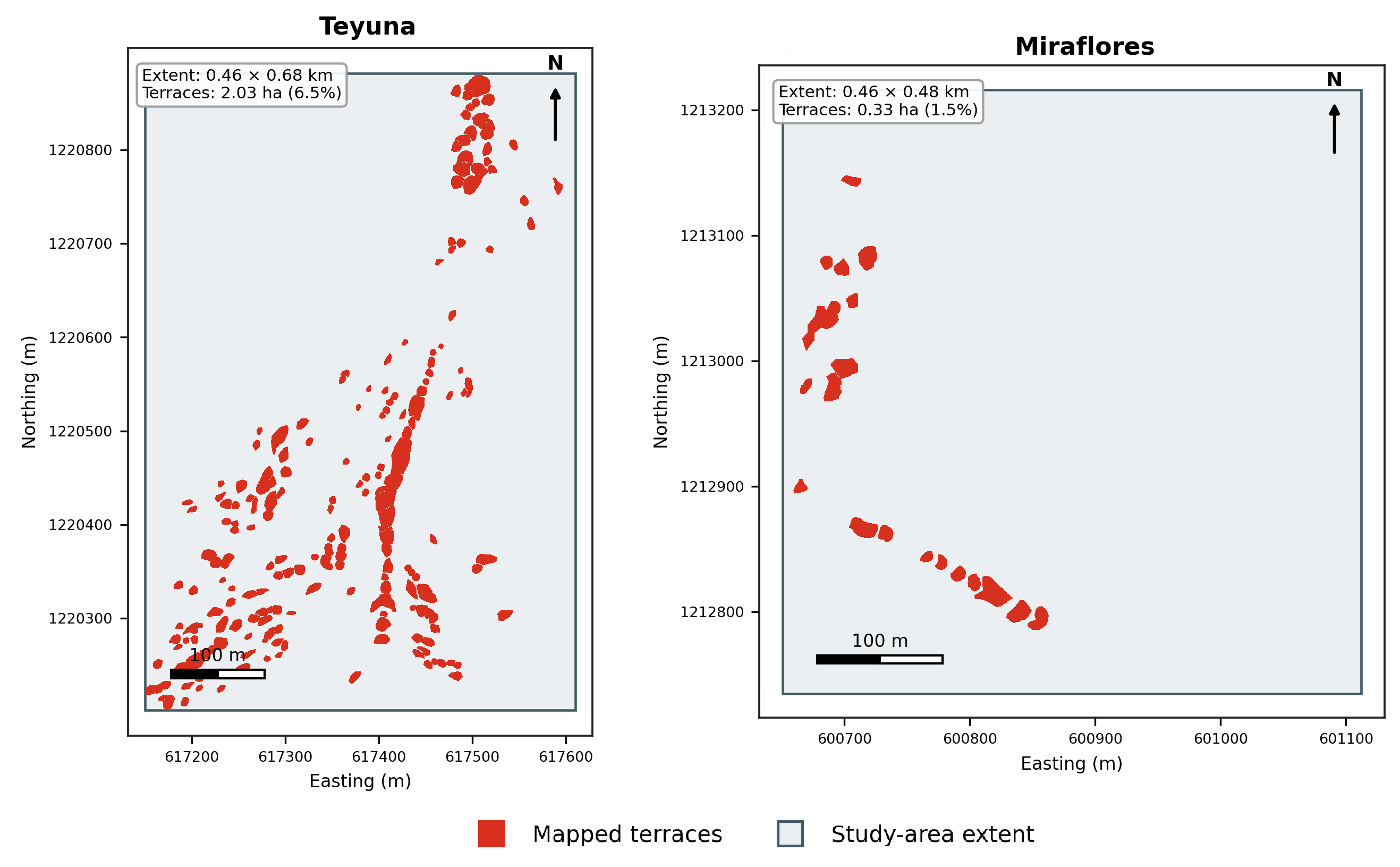}
    \caption{Reference terrace annotations for the two study regions.
    Left: Teyuna (Ciudad Perdida); right: Miraflores.
    Reference terrace annotations are shown over the labelled study extents.
    Mapped terraces cover approximately $6.5\%$ of the labelled Teyuna
    extent and approximately $1.5\%$ of the Miraflores extent.}
    \label{fig:study_area}
\end{figure}

The experiments use two study regions in the Sierra Nevada de Santa Marta, northern Colombia: \textbf{Teyuna} (Ciudad Perdida; approximately $11.04^{\circ}$N, $73.93^{\circ}$W) and \textbf{Miraflores} (approximately $10.97^{\circ}$N, $74.08^{\circ}$W), situated roughly $18\,\mathrm{km}$ apart (Figure~\ref{fig:study_area}). Two transfer directions are defined. In \textbf{Teyuna~$\rightarrow$~Miraflores}, labeled point cloud tiles from Teyuna form the source domain and Miraflores is the unlabeled target/test domain; in \textbf{Miraflores~$\rightarrow$~Teyuna}, the roles are reversed. The two regions differ markedly in terrace density: mapped terraces cover approximately $2.0\,\mathrm{ha}$ ($\approx6.5\%$) of the labelled Teyuna extent but only $\approx0.33\,\mathrm{ha}$ ($\approx1.5\%$) of the Miraflores extent. The two regions also differ in the preservation and present-day morphology of the terraces. In Teyuna, many terraces within the archaeological site have been restored and maintained, resulting in comparatively well-defined terrace forms. In Miraflores, the mapped terraces have not undergone comparable restoration and are generally expressed as more degraded landscape features. Consequently, the cross-region shift considered here involves not only differences in LiDAR acquisition and point-cloud characteristics, but also differences in archaeological feature morphology, preservation state, prevalence, and surrounding landscape context.

\subsection{Lidar acquisition}

We collected lidar data in Miraflores using the YellowScan Mapper+, an UAV LiDAR system based on a Livox Avia laser unit with a pulse repetition frequency (PRF) of 240 kHz. The system was mounted on a DJI Matrice 300 RTK, and we conducted six overlapping flights at various angles and route orientations, with altitudes between 40 and 60 m. This acquisition obtained an approximate all-return density of 2000 $points/m^2$ in triple-echo mode. Data were acquired in August 2024 with dense canopy but sparse understorey foliage. These conditions yield ground-return mean density of 147 $points/m^2$  (SD = 325). The reported vertical accuracy is 3-5 cm. Although the ground return density heterogenous, the ground penetration of the sensor is spatially well distributed, with very often 10 $points/m^2$ across the study area.  

The Teyuna dataset was instead flown from a crewed platform in March 2019 by the National Geographic Society \citep{osorio2023beyond}, using a Riegl VUX-1LR scanner in Multiple-Time-Around mode at a pulse rate of 820 kHz, carried by a Bell 206L3 Long Ranger helicopter cruising at 152 m and 20.6 m/s. Trajectory and attitude were recorded by an Applanix AV410 IMU logging at 200 Hz, while the scanner operated across a 110° field of view with 50\% sidelap between adjacent swaths. The survey window coincided with the close of the dry season in northern Colombia, when canopy and understorey foliage are at their sparsest. These conditions gave an all-return density of 298 $points/m^2$. Reported vertical accuracy for this acquisition is 5–30 cm. The bare-earth subset retains 25 $points/m^2$ on average. The obtained density and its centimetre-scale vertical accuracy in both is sufficient to resolve metre-scale terraces. 

\subsection{Reference annotations}
Terrace reference labels were digitised by field archaeologists on 1 m resolution DTMs derived from the point cloud, and verified in the field. The reference mapping is available as a binary raster at $0.3\,\mathrm{m}$ resolution for each region (Figure~\ref{fig:study_area}), and each point carries a binary label, terrace versus non-terrace. Mapped terraces constitute $\approx6.5\%$ of the labelled Teyuna extent and $\approx1.5\%$ of the Miraflores extent by area, so the task is strongly class-imbalanced. The target/test tiles are treated as unlabelled during adaptation; their labels are used only for the final evaluation reported in Section~\ref{sec:results}.

\subsection{Tiling, features, and preprocessing}

Each domain is partitioned into non-overlapping spatial tiles covering $20\,\mathrm{m}\times20\,\mathrm{m}$. Each input point is represented initially by its 3D coordinate $(x,y,z)$. Hereafter, \emph{tile} refers to the fixed $20\,\mathrm{m}\times20\,\mathrm{m}$ spatial unit, whereas \emph{patch} refers to the fixed-size point cloud sample supplied to the network. Following the 4096-point sampling convention used for semantic segmentation in PointNet \citep{qi2017pointnet}, we set the patch size to $P=4096$ points. This fixed cardinality maintains a computationally tractable input while retaining dense local geometric sampling and ensures consistent input dimensions across tiles with different point densities. During training, each tile is randomly subsampled without replacement when it contains at least $P=4096$ points, or resampled with replacement to $P=4096$ points when it contains fewer than $4096$ points, to form a $4096$-point input patch. The sampled patch coordinates are then recentred to the patch centroid and isotropically scaled so that the most distant point lies on the unit sphere, removing absolute position and gross scale while preserving local geometry. For the KNN-geometry experiments, six local descriptors are computed from the normalised patch coordinates and concatenated with the normalised $(x,y,z)$ coordinates. These descriptors are the estimated surface normal $(n_x,n_y,n_z)$, the slope $s$, and $\cos(a)$ and $\sin(a)$ of the aspect $a$ (Section~\ref{sec:knn}). This yields a nine-channel input, while the coordinate-only configuration remains three-channel. To evaluate over an entire study region without changing the network input size, every test tile is partitioned into deterministic, non-overlapping chunks of $4096$ points; each chunk is processed as one input patch, only the final incomplete chunk is padded, and predictions are accumulated so that every original point receives exactly one prediction (Section~\ref{sec:inference}).

\section{MELD framework}
\label{sec:method}

\begin{figure*}[!htbp]
    \centering
    \includegraphics[width=0.88\textwidth]{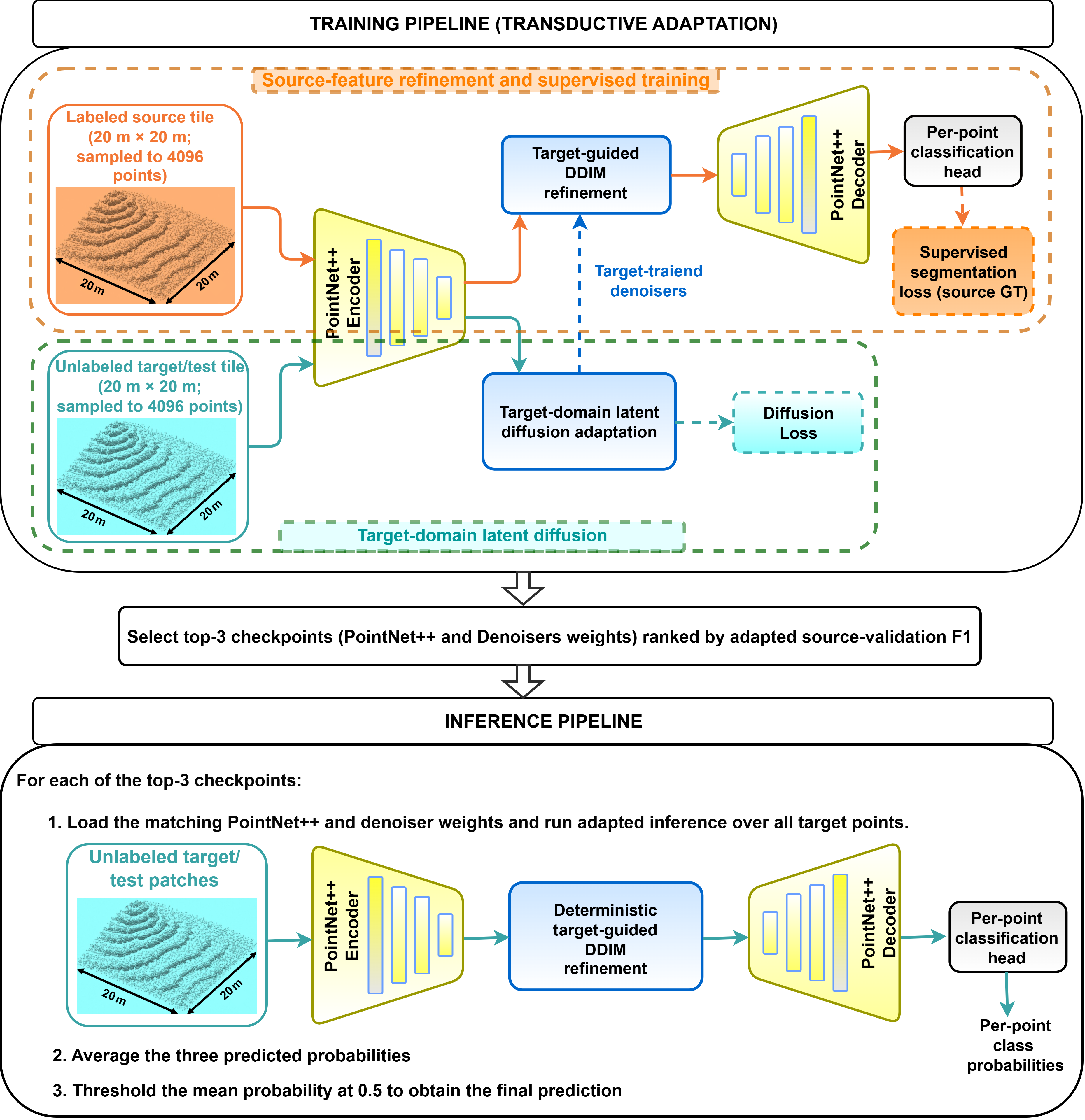}
    \caption{High-level overview of the MELD framework. During joint training, unlabeled target/test features train the level-specific diffusion denoisers, while labeled source features are refined by the target-trained denoisers and decoded under source ground-truth supervision. The diffusion loss updates the denoisers, whereas the segmentation loss updates PointNet++ through the straight-through refinement path. A held-out source-validation set ranks checkpoints by adapted validation F1. At inference, the top three paired PointNet++ denoiser checkpoints are applied independently to all target/test points using deterministic target-guided DDIM refinement; their terrace-class probabilities are averaged point-wise and thresholded at 0.5 to obtain the final binary prediction. No target labels are used for adaptation or checkpoint selection. Preprocessing and the PointNet++ backbone are detailed in Figure~\ref{fig:pointnet_backbone}, target domain latent diffusion adaptation in Figure~\ref{fig:target_diffusion}, and source-feature refinement with the straight-through estimator in Figure~\ref{fig:source_refinement}; adapted inference is described in Section~\ref{sec:inference}.}
    \label{fig:method}
\end{figure*}

Figure~\ref{fig:method} provides a high-level view of the joint transductive training and adapted inference pipelines. Labeled source data provide segmentation supervision, whereas the unlabeled target/test data train level-specific latent denoisers that are subsequently used for source-feature refinement and deterministic adapted inference. Input point cloud patches are denoted by \(X_{\mathrm{src}}\) and \(X_{\mathrm{tar}}\) for the source and target domains, respectively. The same \(\mathrm{src}\) and \(\mathrm{tar}\) subscripts distinguish source and target intermediate representations. Superscript \(l\) indexes the PointNet++ feature level, and \(i\) indexes a flattened latent token.

\subsection{Problem formulation}

Let \(\mathcal{D}_s=\{(X_{\mathrm{src},n},Y_{\mathrm{src},n})\}_{n=1}^{N_s}\) denote labeled source-domain point cloud patches and \(\mathcal{D}_t=\{X_{\mathrm{tar},m}\}_{m=1}^{N_t}\) denote unlabeled target domain point cloud patches, each formed from a spatial tile by the fixed-size point sampling/padding and preprocessing described in Section~\ref{sec:data}. We use the subscripts \(\mathrm{src}\) and \(\mathrm{tar}\) consistently to identify source and target domain inputs and intermediate representations; sample indices are appended when needed. Thus, a generic source input patch is \(X_{\mathrm{src}}\) and a generic target input patch is \(X_{\mathrm{tar}}\). Each patch \(X\in\mathbb{R}^{P\times C}\) contains \(P=4096\) points and \(C\in\{3,9\}\) channels (with or without KNN geometry). The objective is to predict a binary terrace label for each target domain point without using target labels during training or model selection. The protocol is \emph{transductive}: the same unlabeled target \emph{tiles} from which target training patches are sampled are later evaluated exhaustively at test time. The exact 4096-point inputs used during training and inference are not required to be identical: training constructs patches by stochastic sampling from each tile, whereas full-tile inference deterministically partitions the tile points into 4096-point chunks. Target labels are accessed only for final evaluation.

Let \(\mathcal{A}\subseteq\{1,2,3,4\}\) denote the set of active PointNet++ diffusion levels. The principal configuration uses \(\mathcal{A}=\{1,2,3,4\}\), while the level ablations use subsets of this set.

\subsection{PointNet++ segmentation backbone}

The backbone is a PointNet++ encoder--decoder with four set-abstraction (SA) levels and four feature-propagation (FP) levels. Let \(\theta\) denote the trainable parameters of the PointNet++ segmentation network, and let \(f_\theta\) denote its shared encoder component that maps an input patch to the four hierarchical latent feature tensors. The encoder is applied to source and target minibatches using the input-patch notation defined above:
\begin{equation}
\begin{aligned}
    \{Z_{\mathrm{src}}^l\}_{l=1}^{4} &= f_\theta(X_{\mathrm{src}}),
    &\qquad
    \{Z_{\mathrm{tar}}^l\}_{l=1}^{4} &= f_\theta(X_{\mathrm{tar}}),\\
    Z_d^l &\in\mathbb{R}^{B\times C_l\times N_l},
    & d &\in\{\mathrm{src},\mathrm{tar}\}.
\end{aligned}
    \label{eq:latent_tensor}
\end{equation}
where \(B\) is the current minibatch size, \(N_l\) is the number of latent tokens per sample, and \(C_l\) is the feature dimension of each token. At levels \(l=1,2,3,4\), the implemented token counts are \(1024\), \(256\), \(64\), and \(16\), respectively, with corresponding feature dimensions of \(64\), \(128\), \(256\), and \(512\).
A level-\(l\) token is the \(C_l\)-dimensional feature vector associated with one point sampled by the PointNet++ set-abstraction layer, which serves as the center of a local neighbourhood. For a minibatch of size \(B\), flattening a level-\(l\) tensor over the batch and centroid dimensions yields \(M_l=BN_l\) latent tokens. The target and source branch token matrices are defined explicitly in Sections~\ref{sec:diffusion} and~\ref{sec:refine}, respectively. The decoder reverses the hierarchy through FP4--FP1. Level \(4\) provides the deepest FP4 input, while levels \(3\), \(2\), and \(1\) provide skip features to FP4, FP3, and FP2, respectively. FP1 restores the original \(4096\)-point resolution. A point-wise classification head returns two-class log-probabilities. These log-probabilities are used directly by the class-weighted NLL objective during supervised training, whereas exponentiation yields the per-point class probabilities used for adapted inference and ensembling.

\begin{figure*}[!htbp]
    \centering
    \includegraphics[width=0.98\textwidth]{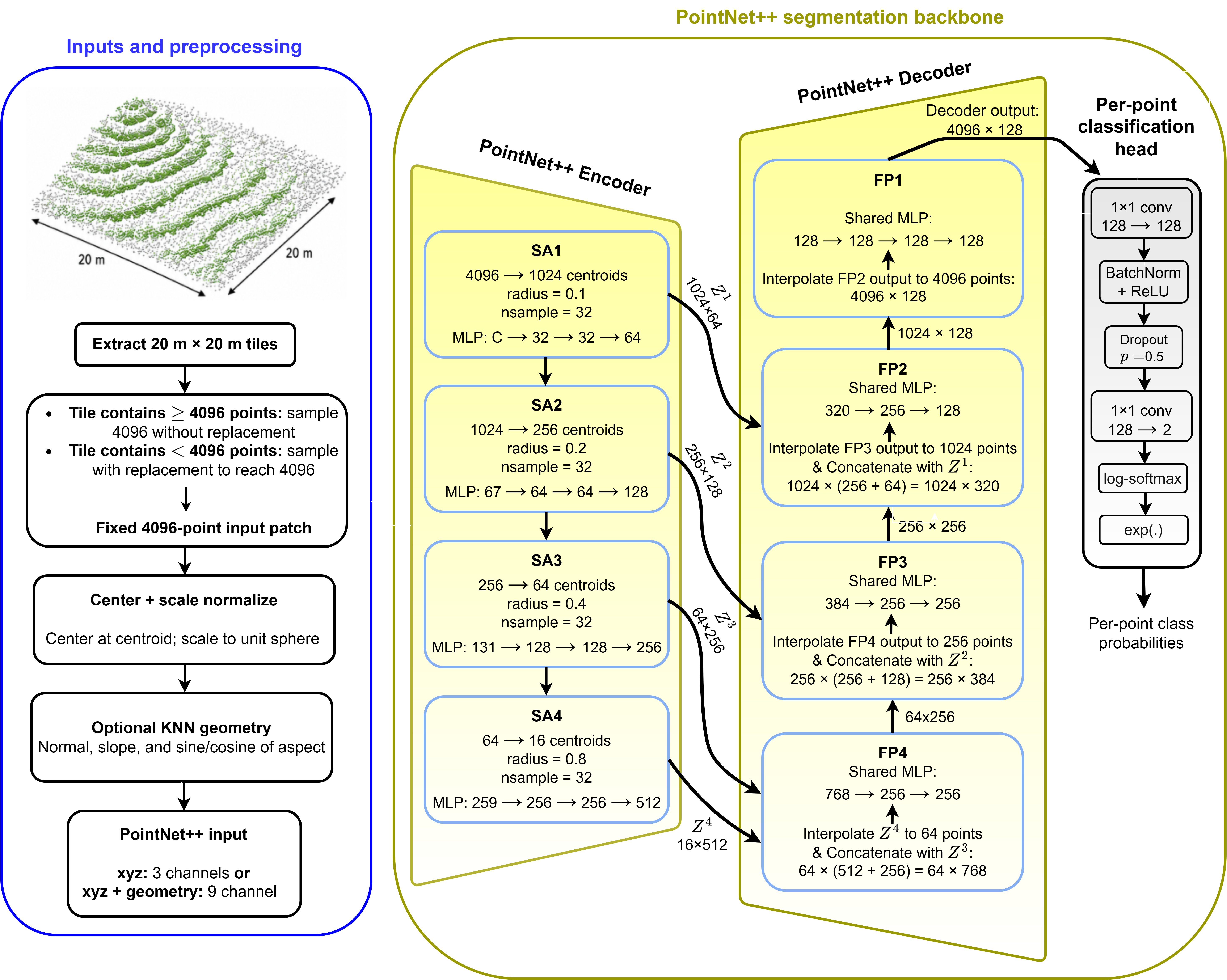}
    \caption{PointNet++ segmentation backbone. Each $20\,\mathrm{m}\times20\,\mathrm{m}$ point cloud tile is converted by sampling or padding into a fixed 4096-point input patch, which is then normalized. The patch may contain either only xyz coordinates (\(C=3\)) or xyz coordinates augmented with optional KNN-derived geometric features (\(C=9\)), including normals, slope, and sine/cosine of aspect. The encoder comprises four set-abstraction (SA) layers that progressively reduce the number of points while increasing feature dimensionality, producing hierarchical latent features \(Z^1\), \(Z^2\), \(Z^3\), and \(Z^4\). The decoder then uses four feature-propagation (FP) stages with interpolation and skip connections to recover point-wise features at the original resolution. A per-point classification head, implemented with \(1\times1\) convolutions, batch normalization, ReLU, dropout, and log-softmax, produces per-point class log-probabilities; applying \(\exp(\cdot)\) yields the corresponding per-point class probabilities. The generic $Z^l$ notation used in this figure denotes the multi-level encoder features that serve as latent representations in the target-adaptation and source-refinement stages, where they are written explicitly as $Z_{\mathrm{tar}}^l$ and $Z_{\mathrm{src}}^l$ in Figures \ref{fig:target_diffusion} and \ref{fig:source_refinement}, respectively.}
    \label{fig:pointnet_backbone}
\end{figure*}

\subsection{target domain latent diffusion}
\label{sec:diffusion}

For each active level \(l\in\mathcal{A}\), a separate token-wise denoiser \(D_l(\cdot,\cdot;\phi_l)\), parameterized by \(\phi_l\), is trained on unlabeled target domain latent tokens. The four denoisers share the same architectural template but have distinct parameters \(\phi_l\) and level-specific input and hidden dimensions. There is no parameter sharing across levels and no interaction between tokens inside a denoiser.

\begin{figure*}[!htbp]
    \centering
    \includegraphics[width=1.07\textwidth]{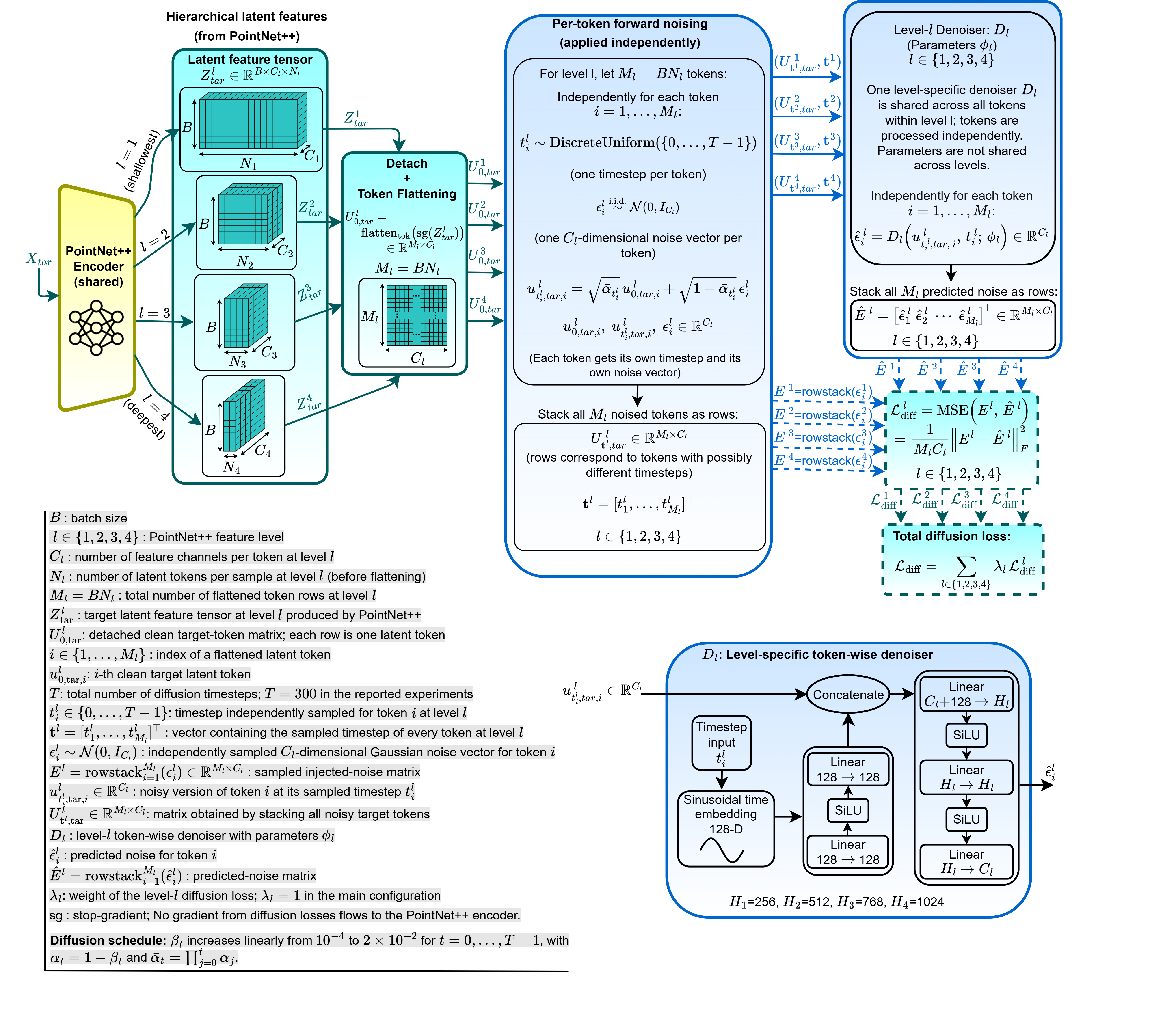}
    \caption{target domain latent diffusion adaptation. Unlabeled target domain input patches \(X_{\mathrm{tar}}\) are passed through the shared PointNet++ encoder to obtain multi-level latent features \(Z_{\mathrm{tar}}^l\), \(l\in\{1,2,3,4\}\). At each level, the target features are detached from the backbone and flattened into token matrices \(U_{0,\mathrm{tar}}^l\), so the target domain diffusion objective does not propagate gradients into PointNet++. Each token is independently assigned a diffusion timestep \(t_i^l\) and perturbed with Gaussian noise according to the forward diffusion process. A level-specific denoiser \(D_l(\cdot,\cdot;\phi_l)\), conditioned on the noisy token and its timestep embedding, predicts the added noise. Within each level, the same denoiser is shared across all tokens, while denoiser parameters are not shared between feature levels. The level-wise noise-prediction losses are computed as mean-squared error between the sampled and predicted noise and are combined into the total diffusion loss \(\mathcal{L}_{\mathrm{diff}}\). Consequently, target domain adaptation updates only the denoiser parameters \(\phi_l\), allowing the denoisers to learn the latent feature distribution of the unlabeled target domain for subsequent source-feature refinement and adapted inference. The figure shows the principal configuration in which all four PointNet++ feature levels are active; the level-ablation experiments use the corresponding subsets of these levels.}
    \label{fig:target_diffusion}
\end{figure*}

The target feature tensor is detached before token flattening so that the diffusion objective does not update the PointNet++ encoder:
\begin{equation}
    U_{0,\mathrm{tar}}^l
    =\operatorname{flatten}_{\mathrm{tok}}\!\left(\operatorname{sg}(Z_{\mathrm{tar}}^l)\right)
    \in\mathbb{R}^{M_l\times C_l},
    \label{eq:target_flatten}
\end{equation}
where \(\operatorname{sg}(\cdot)\) denotes stop-gradient. For each flattened token
\(i\in\{1,\ldots,M_l\}\), a diffusion timestep and a Gaussian noise vector are
sampled independently:
\begin{equation}
    t_i^l \overset{\mathrm{i.i.d.}}{\sim}
    \operatorname{DiscreteUniform}\{0,\ldots,T-1\},
    \qquad
    \epsilon_i^l \overset{\mathrm{i.i.d.}}{\sim}
    \mathcal{N}(0,I_{C_l}).
    \label{eq:token_sampling}
\end{equation}
Here, \(T\) is the total number of diffusion timesteps and \(I_{C_l}\) is the
\(C_l\times C_l\) identity matrix. Thus, each token receives its own independently
sampled timestep \(t_i^l\) and \(C_l\)-dimensional Gaussian noise vector
\(\epsilon_i^l\).

The forward noising operation for token \(i\) is
\begin{equation}
    u_{t_i^l,\mathrm{tar},i}^l
    =
    \sqrt{\bar{\alpha}_{t_i^l}}\,u_{0,\mathrm{tar},i}^l
    +
    \sqrt{1-\bar{\alpha}_{t_i^l}}\,\epsilon_i^l.
    \label{eq:forward_token}
\end{equation}
Here, \(u_{0,\mathrm{tar},i}^l\in\mathbb{R}^{C_l}\) is the \(i\)-th clean
target-domain latent token at PointNet++ feature level \(l\), while
\(u_{t_i^l,\mathrm{tar},i}^l\in\mathbb{R}^{C_l}\) is its noisy version at the
sampled timestep \(t_i^l\). The coefficient \(\bar{\alpha}_{t_i^l}\) determines
the proportion of the original token retained at that timestep and is defined
by the diffusion schedule below. Consequently,
\(\sqrt{\bar{\alpha}_{t_i^l}}\) scales the clean token and
\(\sqrt{1-\bar{\alpha}_{t_i^l}}\) scales the injected Gaussian noise.

The sampled timesteps and noisy tokens are then stacked as
\begin{equation}
    \mathbf{t}^l
    =
    [t_1^l,\ldots,t_{M_l}^l]^\top,
    \qquad
    U_{\mathbf{t}^l,\mathrm{tar}}^l
    =
    \operatorname{rowstack}_{i=1}^{M_l}
    \left(u_{t_i^l,\mathrm{tar},i}^l\right)
    \in\mathbb{R}^{M_l\times C_l}.
    \label{eq:noisy_matrix}
\end{equation}
Because the timestep is sampled independently for each token, different rows of
\(U_{\mathbf{t}^l,\mathrm{tar}}^l\) can correspond to different diffusion
timesteps.

The variance-preserving diffusion schedule is defined for
\(t=0,\ldots,T-1\) by
\begin{equation}
    \beta_t
    =
    \beta_{\min}
    +
    \frac{t}{T-1}
    (\beta_{\max}-\beta_{\min}),
    \qquad
    \beta_{\min}=10^{-4},
    \qquad
    \beta_{\max}=2\times10^{-2},
    \label{eq:beta_schedule}
\end{equation}
\begin{equation}
    \alpha_t=1-\beta_t,
    \qquad
    \bar{\alpha}_t=\prod_{j=0}^{t}\alpha_j,
    \qquad
    T=300.
    \label{eq:alpha_schedule}
\end{equation}

Each denoiser takes a noisy token together with its corresponding diffusion
timestep and predicts the noise that was added:
\begin{equation}
    \widehat{\epsilon}_i^l
    =
    D_l\!\left(
        u_{t_i^l,\mathrm{tar},i}^l,
        t_i^l;
        \phi_l
    \right)
    \in\mathbb{R}^{C_l}.
    \label{eq:denoiser}
\end{equation}
Internally, \(D_l\) maps \(t_i^l\) to a fixed \(128\)-dimensional sinusoidal embedding that is refined by a two-layer SiLU MLP and concatenated with the noisy token. The resulting \((C_l+128)\)-dimensional vector is processed by a three-layer MLP with SiLU activations, hidden width \(H_l\), and a linear output layer of dimension \(C_l\), where \(H_1=256\), \(H_2=512\), \(H_3=768\), and \(H_4=1024\). This architecture is illustrated in Figure~\ref{fig:target_diffusion}. The same \(D_l\) weights are applied independently to every token at level \(l\).

For all \(M_l\) tokens at feature level \(l\) in the minibatch, the sampled and predicted noise vectors are stacked row-wise to form
\begin{equation}
    E^l
    =\operatorname{rowstack}_{i=1}^{M_l}(\epsilon_i^l),
    \qquad
    \widehat{E}^l
    =\operatorname{rowstack}_{i=1}^{M_l}(\widehat{\epsilon}_i^l),
    \qquad
    E^l,\widehat{E}^l\in\mathbb{R}^{M_l\times C_l}.
    \label{eq:noise_matrices}
\end{equation}
Here, \(E^l\) contains the Gaussian noise actually injected into the target tokens, while \(\widehat{E}^l\) contains the corresponding noise predicted by the level-specific denoiser \(D_l\).

The level-\(l\) diffusion loss is the mean-squared error between the sampled and predicted noise over all \(M_lC_l\) token-channel elements:
\begin{equation}
    \mathcal{L}_{\mathrm{diff}}^l
    =
    \operatorname{MSE}(E^l,\widehat{E}^l)
    =
    \frac{1}{M_lC_l}
    \left\|E^l-\widehat{E}^l\right\|_F^2.
    \label{eq:ldiff}
\end{equation}

The diffusion losses from all active feature levels are combined as
\begin{equation}
    \mathcal{L}_{\mathrm{diff}}
    =
    \sum_{l\in\mathcal{A}}
    \lambda_l\mathcal{L}_{\mathrm{diff}}^l,
    \label{eq:total_diff}
\end{equation}
where \(\lambda_l\) controls the contribution of level \(l\) and is set to \(1\) for every active level in the principal configuration. Because the target features are detached in Eq.~\eqref{eq:target_flatten}, gradients from \(\mathcal{L}_{\mathrm{diff}}\) update only the denoiser parameters \(\phi_l\) and do not propagate into the PointNet++ backbone.

\subsection{Source-feature refinement and supervised training}
\label{sec:refine}

\begin{figure*}[!htbp]
    \centering
    \includegraphics[width=1\textwidth]{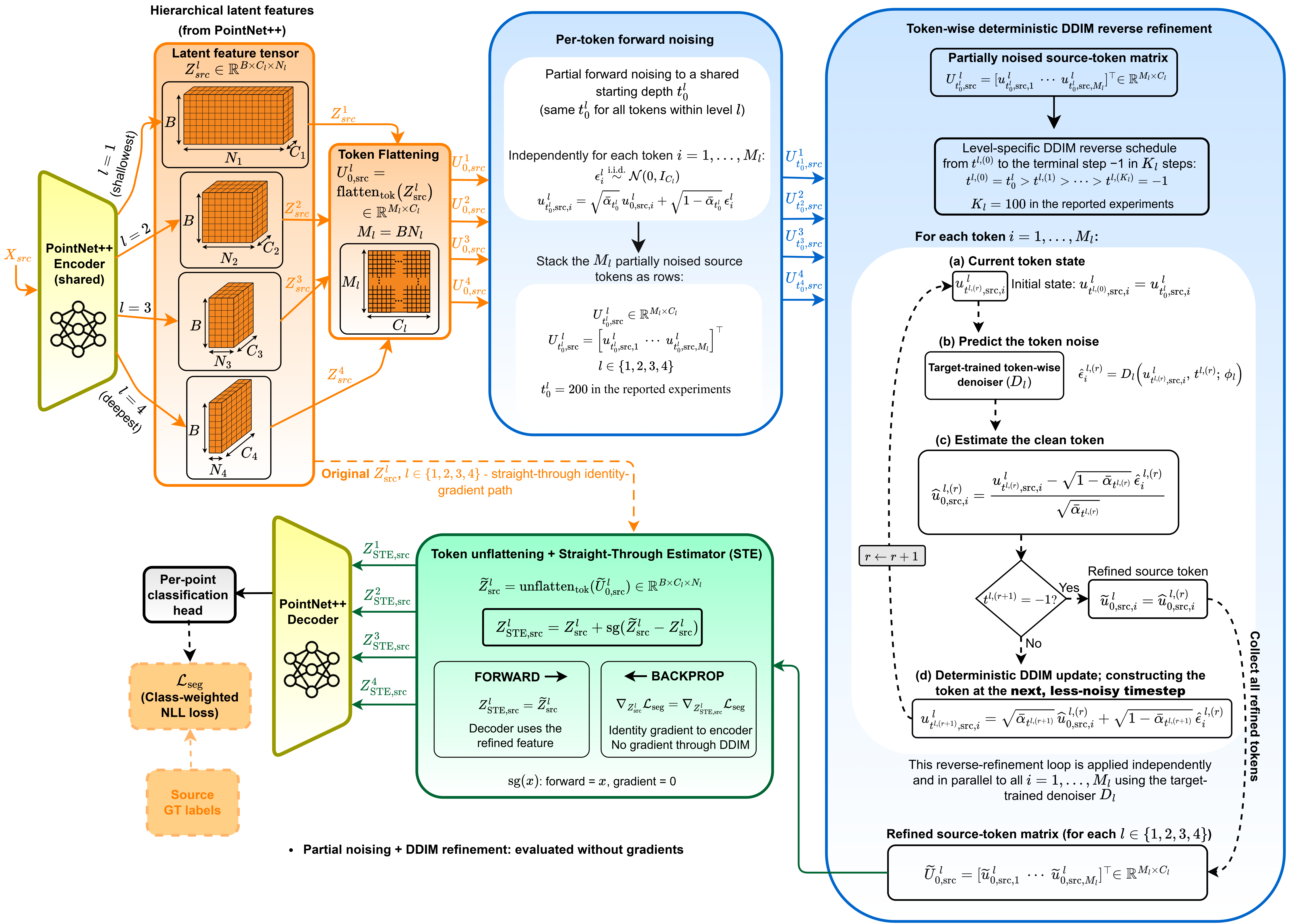}
    \caption{Source refinement for supervised segmentation training. A labeled source-domain input patch \(X_{\mathrm{src}}\) is passed through the shared PointNet++ encoder to obtain hierarchical latent features \(Z_{\mathrm{src}}^l\) for \(l\in\{1,2,3,4\}\). At each feature level, the encoder output is tokenized into \(U_{0,\mathrm{src}}^l\), partially noised to a shared starting depth \(t_0^l\), and then refined token-wise through a deterministic DDIM reverse process using the level-specific target-trained denoiser \(D_l\). The refined token matrix \(\widetilde{U}_{0,\mathrm{src}}^l\) is unflattened to \(\widetilde{Z}_{\mathrm{src}}^l\) and supplied to the decoder through a straight-through estimator (STE), so that the decoder uses the refined features in the forward pass while segmentation gradients flow back to the original encoder features during backpropagation. The decoder output is supervised by the source ground-truth labels using the class-weighted NLL loss \(\mathcal{L}_{\mathrm{seg}}\). The figure shows the principal configuration in which all four PointNet++ feature levels are refined. During training, source refinement is stochastic overall because the initial Gaussian perturbation is sampled randomly, whereas the subsequent DDIM reverse trajectory is deterministic conditional on the resulting partially noised token.}
    \label{fig:source_refinement}
\end{figure*}

In the source branch of the same joint training procedure, the level-specific denoisers trained through the target domain diffusion objective are used to refine source features before supervised decoding (Figure~\ref{fig:source_refinement}).

For each active level \(l\), source features are flattened as
\begin{equation}
    U_{0,\mathrm{src}}^l
    =\operatorname{flatten}_{\mathrm{tok}}(Z_{\mathrm{src}}^l)
    =\operatorname{rowstack}_{i=1}^{M_l}(u_{0,\mathrm{src},i}^l)
    \in\mathbb{R}^{M_l\times C_l}.
    \label{eq:source_flatten}
\end{equation}
The refinement branch itself is evaluated without gradients. During training, all tokens within a level use the same starting depth \(t_0^l\), but each token receives an independent Gaussian perturbation,
\begin{equation}
    \epsilon_i^l
    \overset{\mathrm{i.i.d.}}{\sim}\mathcal{N}(0,I_{C_l}),
    \qquad
    u_{t_0^l,\mathrm{src},i}^l
    =\sqrt{\bar{\alpha}_{t_0^l}}\,u_{0,\mathrm{src},i}^l
    +\sqrt{1-\bar{\alpha}_{t_0^l}}\,\epsilon_i^l.
    \label{eq:source_partial_noise}
\end{equation}
The partially noised source tokens are then stacked into the level-\(l\) matrix
\begin{equation}
    U_{t_0^l,\mathrm{src}}^l
    =\operatorname{rowstack}_{i=1}^{M_l}\!\left(u_{t_0^l,\mathrm{src},i}^l\right)
    \in\mathbb{R}^{M_l\times C_l},
    \label{eq:source_noisy_matrix}
\end{equation}
which provides the initial states for the token-wise DDIM reverse trajectories. In the principal training configuration, \(t_0^l=200\) for all active levels.

A level-specific reverse schedule contains \(K_l\) deterministic DDIM updates,
\begin{equation}
    t^{l,(0)}=t_0^l>t^{l,(1)}>\cdots>t^{l,(K_l-1)}\geq0,
    \qquad
    t^{l,(K_l)}=-1,
    \label{eq:ddim_schedule}
\end{equation}
where \(-1\) is a termination marker indicating that the final clean-token estimate is returned. No diffusion coefficient is evaluated at \(t=-1\). The reported training configuration uses \(K_l=100\) at every active level.

At reverse iteration \(r\), given the current token state \(u_{t^{l,(r)},\mathrm{src},i}^l\), the level-specific denoiser predicts
\begin{equation}
    \widehat{\epsilon}_i^{l,(r)}
    =
    D_l\!\left(
        u_{t^{l,(r)},\mathrm{src},i}^l,
        t^{l,(r)};
        \phi_l
    \right).
    \label{eq:ddim_eps}
\end{equation}
The corresponding clean-token estimate is
\begin{equation}
    \hat{u}_{0,\mathrm{src},i}^{l,(r)}
    =
    \frac{
        u_{t^{l,(r)},\mathrm{src},i}^l
        -
        \sqrt{1-\bar{\alpha}_{t^{l,(r)}}}\,
        \widehat{\epsilon}_i^{l,(r)}
    }{
        \sqrt{\bar{\alpha}_{t^{l,(r)}}}
    }.
    \label{eq:ddim_x0}
\end{equation}

If \(t^{l,(r+1)}\geq0\), the deterministic DDIM update computes the token at the next, less-noisy timestep as
\begin{equation}
    u_{t^{l,(r+1)},\mathrm{src},i}^l
    =
    \sqrt{\bar{\alpha}_{t^{l,(r+1)}}}\,
    \hat{u}_{0,\mathrm{src},i}^{l,(r)}
    +
    \sqrt{1-\bar{\alpha}_{t^{l,(r+1)}}}\,
    \widehat{\epsilon}_i^{l,(r)}.
    \label{eq:ddim_update}
\end{equation}

If \(t^{l,(r+1)}=-1\), the trajectory terminates with
\begin{equation}
    \tilde{u}_{0,\mathrm{src},i}^l
    =
    \hat{u}_{0,\mathrm{src},i}^{l,(r)}.
    \label{eq:ddim_terminal}
\end{equation}

No fresh random noise is introduced during the reverse trajectory. Therefore, for a fixed partially noised starting token, Eqs.~\eqref{eq:ddim_eps}--\eqref{eq:ddim_terminal} define a deterministic refinement trajectory. Source refinement during training is nevertheless stochastic overall because the initial perturbation in Eq.~\eqref{eq:source_partial_noise} is random.

The refined tokens are stacked and reshaped back to the PointNet++ tensor layout,
\begin{equation}
    \tilde{U}_{0,\mathrm{src}}^l
    =\operatorname{rowstack}_{i=1}^{M_l}(\tilde{u}_{0,\mathrm{src},i}^l)
    \in\mathbb{R}^{M_l\times C_l},
    \qquad
    \tilde{Z}_{\mathrm{src}}^l
    =\operatorname{unflatten}_{\mathrm{tok}}(\tilde{U}_{0,\mathrm{src}}^l)
    \in\mathbb{R}^{B\times C_l\times N_l}.
    \label{eq:source_unflatten}
\end{equation}
These refined features are supplied to the decoder through a straight-through estimator (STE):
\begin{equation}
    Z_{\mathrm{STE},\mathrm{src}}^l
    =Z_{\mathrm{src}}^l
    +\operatorname{sg}\!\left(\tilde{Z}_{\mathrm{src}}^l-Z_{\mathrm{src}}^l\right).
    \label{eq:ste}
\end{equation}
The STE therefore satisfies
\begin{equation}
\begin{aligned}
    Z_{\mathrm{STE},\mathrm{src}}^l
    &=\tilde{Z}_{\mathrm{src}}^l,
    && \text{in the forward pass},\\
    \frac{\partial Z_{\mathrm{STE},\mathrm{src}}^l}
         {\partial Z_{\mathrm{src}}^l}
    &= I,
    && \text{in the backward pass}.
\end{aligned}
\label{eq:ste_properties}
\end{equation}
Thus, the decoder receives the refined feature values, while the segmentation gradient passes unchanged to the original PointNet++ encoder features and does not propagate through the noising or DDIM refinement trajectory. The refined level-\(4\) feature is used as the deepest FP4 input, while refined levels \(3\), \(2\), and \(1\) replace the corresponding skip features entering FP4, FP3, and FP2, respectively.

Let
\(p_{\theta,bp}(c\mid X_{\mathrm{src}},\{Z_{\mathrm{STE},\mathrm{src}}^l\}_{l\in\mathcal A})\)
denote the predicted probability of class \(c\in\{0,1\}\) for point \(p\) of source sample \(b\), and let \(y_{\mathrm{src},bp}\) denote its ground-truth class. With source-derived class weights \(w_c\), the class-weighted negative log-likelihood is
\begin{equation}
    \mathcal{L}_{\mathrm{seg}}
    =-
    \frac{
      \sum_{b,p} w_{y_{\mathrm{src},bp}}
      \log p_{\theta,bp}\!\left(
      y_{\mathrm{src},bp}
      \mid
      X_{\mathrm{src}},
      \{Z_{\mathrm{STE},\mathrm{src}}^l\}_{l\in\mathcal A}
      \right)
    }{
      \sum_{b,p} w_{y_{\mathrm{src},bp}}
    }.
    \label{eq:seg_loss}
\end{equation}

The complete training objective is
\begin{equation}
    \mathcal{L}_{\mathrm{total}}
    =\mathcal{L}_{\mathrm{seg}}
    +\mathcal{L}_{\mathrm{diff}}.
    \label{eq:total_loss}
\end{equation}
Because the target features are detached in Eq.~\eqref{eq:target_flatten} and source refinement is evaluated without gradients before applying the STE, the gradient contributions to the PointNet++ and denoiser parameters are separated:
\begin{equation}
    \nabla_\theta\mathcal{L}_{\mathrm{total}}
    =\nabla_\theta\mathcal{L}_{\mathrm{seg}},
    \qquad
    \nabla_{\phi_l}\mathcal{L}_{\mathrm{total}}
    =\lambda_l\nabla_{\phi_l}\mathcal{L}_{\mathrm{diff}}^l.
    \label{eq:gradient_routing}
\end{equation}
Thus, the segmentation objective trains the PointNet++ encoder, decoder, and classification head, while each target-domain diffusion loss trains only its corresponding level-specific denoiser \(D_l\).

\subsection{Adapted inference and ensembling}
\label{sec:inference}

Adapted validation and target/test inference reuse the same level-specific denoisers and DDIM reverse equations, but initialize the evaluation trajectory with a zero Gaussian-noise tensor to obtain deterministic and reproducible feature refinement. This removes variability due to random noise sampling, so that repeated evaluation of the same feature tensor with the same checkpoint produces the same refinement trajectory. For a validation or target/test token \(u_{0,d,i}^{l}\), where \(d\in\{\mathrm{val},\mathrm{tar}\}\) denotes the source-validation or target/test split, respectively, the evaluation trajectory starts at
\begin{equation}
    u_{t_{0,\mathrm{eval}}^{l},d,i}^{l}
    =
    \sqrt{\bar{\alpha}_{t_{0,\mathrm{eval}}^{l}}}\,
    u_{0,d,i}^{l},
    \label{eq:eval_initial}
\end{equation}
which follows from the forward diffusion equation when the Gaussian perturbation is set to zero. The reported configuration uses \(t_{0,\mathrm{eval}}^{l}=100\) and \(K_{l,\mathrm{eval}}=50\) reverse DDIM updates at every active level. Although no random Gaussian perturbation is added at initialization, the clean token is first mapped to the selected diffusion timestep as
\(u_t=\sqrt{\bar{\alpha}_t}\,u_0\).
The target-trained denoiser then estimates a correction from this timestep-dependent latent state. Because this predicted correction is not constrained to zero, the subsequent DDIM reverse updates can modify the latent representation throughout the refinement trajectory. The resulting refined token matrices are unflattened to \(\widetilde{Z}_{d}^{l}\) and supplied directly to the corresponding decoder pathways. No STE is required during validation or inference because no backward pass is performed.

During each run, the three checkpoints with the highest \emph{adapted source-validation} F1 are retained, which are then used for ensemble inference. If \(p_i^{(m)}\) is the terrace probability assigned to point \(i\) by the \(m\)-th selected checkpoint, the final ensemble probability is
\begin{equation}
    P_i=\frac{1}{3}\sum_{m=1}^{3}p_i^{(m)}.
    \label{eq:ensemble}
\end{equation}
Because validation tiles are drawn from the source domain and target labels are never used for checkpoint selection, the ranking procedure does not leak target-label information. For full-region prediction, each test tile is processed using the deterministic \(4096\)-point chunking procedure described in Section~\ref{sec:data}. Each chunk forms one network input patch, ensuring that the inference input size matches that used during training and that every original point receives exactly one prediction.

\subsection{KNN geometric descriptors}
\label{sec:knn}

As an optional complement to the learned representations, each fixed-size
input patch can be augmented with six local geometric descriptors. For a
normalized 3D point $p_q \in \mathbb{R}^3$ in a patch, let
$\mathcal{N}_k(q)$ denote the $k$ points returned by the nearest-neighbour
query for $p_q$, including $p_q$ itself in the implementation, and let
$\bar{p}_q = k^{-1}\sum_{p_r \in \mathcal{N}_k(q)} p_r$ be the neighbourhood
centroid. The local scatter matrix is

\begin{equation}
\Sigma_q =
\sum_{p_r \in \mathcal{N}_k(q)}
(p_r-\bar{p}_q)(p_r-\bar{p}_q)^{\top}.
\label{eq:knn_scatter}
\end{equation}

The unit eigenvector of $\Sigma_q$ associated with its smallest eigenvalue
is used as the local surface normal
$n_q=(n_{x,q},n_{y,q},n_{z,q})^{\top}$. Its sign is flipped when necessary
so that $n_{z,q} \geq 0$, giving a consistent upward orientation. With this
convention, the local slope and orientation angle are defined as

\begin{equation}
s_q = \arccos(n_{z,q}),
\qquad
a_q = \operatorname{atan2}(-n_{y,q},-n_{x,q}).
\label{eq:knn_slope_aspect}
\end{equation}

Here, \(a_q\) is the orientation angle of the negative horizontal projection of the upward-oriented normal, measured from the positive \(x\)-axis according to the \(\operatorname{atan2}\) convention. For brevity, \(a_q\) is referred to as the aspect elsewhere in the paper, with its convention fixed by Eq.~\eqref{eq:knn_slope_aspect}. Because aspect is circular, it is
represented by its cosine and sine rather than by the angle itself. The
resulting six-dimensional geometric descriptor is

\begin{equation}
g_q =
\left[
n_{x,q},
n_{y,q},
n_{z,q},
s_q,
\cos(a_q),
\sin(a_q)
\right]^{\top}
\in \mathbb{R}^{6}.
\label{eq:knn_descriptor}
\end{equation}

These descriptors are computed from the normalized coordinates within each
4096-point input patch using $k=20$ and concatenated with the normalized
$(x,y,z)$ coordinates, yielding the nine-channel input used in the
KNN-geometry experiments. Translation and positive isotropic scaling do not
change the estimated surface orientations or nearest-neighbour ordering.

\section{Experimental setup}
\label{sec:experiments}

\subsection{Data partitions and transductive protocol}

For each transfer direction, the source region provides labeled training tiles and a held-out labeled source-validation set, whereas the target region provides the target/test tiles, which are used without labels during diffusion-denoiser training. Their labels are withheld throughout training, adaptation, and model selection and are accessed only for final evaluation. Model selection is based exclusively on adapted source-validation F1.

\subsection{Training protocol}

All experiments use $4096$ points per patch and a batch size of $16$. Optimisation uses Adam with learning rate $10^{-3}$ for both the backbone and the denoisers and weight decay $10^{-4}$. Models are trained for $200$ epochs. Class weights are computed by inverse frequency from the source training labels. The diffusion horizon is $T=300$ with a linear $\beta$ schedule (Section~\ref{sec:diffusion}). For every active level, $\lambda_l=1$, $t_0^l=200$, and $K_l=100$ during source refinement. Adapted validation and inference use the DDIM starting depth $t_{0,\mathrm{eval}}^l=100$ and $K_{l,\mathrm{eval}}=50$ reverse updates. For numerical stability in the implementation, $10^{-8}$ is added to the denominator of each clean-token estimate (Eq. \ref{eq:ddim_x0}) and the resulting estimate is clipped element-wise to $[-3,3]$ before the subsequent DDIM update, preventing extreme latent values during reverse refinement. Results are reported as mean and standard deviation over three random seeds ($100$, $200$, $300$). 

\subsection{Baselines}
\label{sec:baselines}

We compare MELD against a \textbf{source-only} PointNet++ model (no adaptation), which serves as the primary reference baseline, as well as five established UDA methods: maximum-mean-discrepancy alignment as used in deep adaptation networks (\textbf{MMD/DAN}) \citep{long2015dan}; correlation alignment (\textbf{Deep CORAL}) \citep{sun2016deepcoral}; domain-adversarial training (\textbf{DANN}) \citep{ganin2016dann}; conditional adversarial adaptation with entropy conditioning (\textbf{CDAN+E}) \citep{long2018cdan}; and entropy minimisation on the target predictions (\textbf{MinEnt}) \citep{grandvalet2004}. To make the comparison fair with respect to \emph{where} adaptation acts, the methods that can operate on intermediate features are evaluated in two settings. In the \emph{single} setting, adaptation is applied only at the deepest PointNet++ level ($l=4$), corresponding to the original configuration of these methods. In the \emph{multi} setting, adaptation is applied simultaneously at all four PointNet++ levels ($l\in\{1,2,3,4\}$), matching the levels refined by MELD. This yields nine competing variants in total. MinEnt acts on the output probabilities and therefore has no level variants.

Deep CORAL aligns second-order token statistics independently at each active feature level, using up to $4096$ sampled tokens. DANN and CDAN+E attach a domain discriminator with gradient reversal to each active level. For CDAN+E, the discriminator is conditioned on the outer product of the features and class predictions, with entropy-based weighting. In the \emph{multi} setting, the domain adaptation losses are averaged across the active levels. The adaptation-loss weight is $1.0$ for the feature-level methods and $0.01$ for MinEnt. All baselines use coordinate-only inputs (no KNN geometry) and share the same PointNet++ backbone used by MELD, the same three seeds, the same $200$-epoch schedule and batch size, checkpoint selection by source-validation F1, top-three checkpoint ensembling, and the identical all-points evaluation (Section~\ref{sec:metrics}), so performance differences reflect the adaptation mechanism rather than incidental settings. 

In addition, A target-only oracle provides an in-domain reference for each transfer direction. Specifically, a supervised model with the same backbone and coordinate-only inputs is trained directly on labels from the region that serves as the target domain in that transfer direction, using five-fold cross-validation with three random seeds. In each fold, predictions are generated only for held-out tiles, ensuring that no tile is evaluated by a model trained on that tile's labels. The oracle indicates what in-domain supervision achieves on each region and is reported as a reference point for the absolute scores rather than as a competing UDA method.

\subsection{Evaluation metrics}
\label{sec:metrics}

For true positives $TP$, false positives $FP$, and false negatives $FN$, precision, recall, and F1 are
\begin{equation}
    \mathrm{P}=\frac{TP}{TP+FP},\qquad
    \mathrm{R}=\frac{TP}{TP+FN},\qquad
    \mathrm{F1}=\frac{2TP}{2TP+FP+FN}.
\end{equation}
F1 is the primary metric throughout and is computed at probability threshold $0.5$ from the top-three ensemble-mean probability over \emph{every} point of every evaluation tile (Section~\ref{sec:inference}). Overall accuracy was not used because the task is strongly class-imbalanced, such that this metric would be dominated by the much more abundant non-terrace points and would therefore provide limited diagnostic value. In addition to the primary classification metrics, we report F1--threshold, precision--recall, and ROC curves to examine performance across a range of probability thresholds.

All configurations evaluated in this study, including MELD and its geometry-enhanced variant, all nine baseline variants, and the target-only oracle, are assessed using the same all-points evaluation pipeline. Specifically, they use the same deterministic chunking procedure, with the chunk ordering seeded by the corresponding run seed, the same per-chunk normalisation, the same top-three probability ensembling, and the same decision threshold. Consequently, differences in reported performance cannot be attributed to differences in the evaluation protocol across methods.

\section{Results}
\label{sec:results}

\subsection{Comparison with UDA baselines}
\label{sec:results_main}

Table~\ref{tab:main_results} reports the point-level comparison across both transfer directions: the source-only lower bound, the nine competing UDA variants, MELD, MELD + geometry, and the target-only oracle as an in-domain reference.

\begin{table*}[t]
\centering
\begin{threeparttable}
\caption{Cross-region terrace detection (point level, all points): F1 as mean $\pm$ standard deviation over three random seeds. \emph{Single} applies adaptation at the deepest level ($l=4$); \emph{multi} at all four levels ($l\in\{1,2,3,4\}$). Mean is the unweighted average of the two directions. Best result per column in bold. The target-only oracle (italic) is an in-domain supervised reference trained with five-fold cross-validation on the evaluation region of each direction; it is not a competing UDA method and is reported for context only.}
\label{tab:main_results}
\footnotesize
\setlength{\tabcolsep}{3pt}
\begin{tabular}{@{}llccc@{}}
\toprule
\textbf{Method} & \textbf{Levels} & \textbf{\begin{tabular}[c]{@{}c@{}}Teyuna $\rightarrow$\\ Miraflores\end{tabular}} & \textbf{\begin{tabular}[c]{@{}c@{}}Miraflores $\rightarrow$\\ Teyuna\end{tabular}} & \textbf{Mean} \\
\midrule
Source-only PointNet++ & --- & $0.397 \pm 0.037$ & $0.557 \pm 0.011$ & $0.477$ \\
\addlinespace[2pt]
MMD/DAN & single & $0.422 \pm 0.010$ & $0.540 \pm 0.018$ & $0.481$ \\
MMD/DAN & multi & $0.352 \pm 0.028$ & $0.531 \pm 0.016$ & $0.442$ \\
Deep CORAL & single & $0.342 \pm 0.037$ & $0.572 \pm 0.008$ & $0.457$ \\
Deep CORAL & multi & $0.382 \pm 0.034$ & $0.573 \pm 0.027$ & $0.477$ \\
DANN & single & $0.395 \pm 0.102$ & $0.551 \pm 0.016$ & $0.473$ \\
DANN & multi & $0.385 \pm 0.031$ & $0.585 \pm 0.017$ & $0.485$ \\
CDAN+E & single & $0.363 \pm 0.060$ & $0.563 \pm 0.011$ & $0.463$ \\
CDAN+E & multi & $0.387 \pm 0.046$ & $0.571 \pm 0.029$ & $0.479$ \\
MinEnt & output & $0.405 \pm 0.051$ & $0.574 \pm 0.026$ & $0.489$ \\
\addlinespace[2pt]
MELD & multi & $0.459 \pm 0.038$ & $0.594 \pm 0.015$ & $0.527$ \\
MELD + geometry & multi & $\mathbf{0.542 \pm 0.026}$ & $\mathbf{0.606 \pm 0.032}$ & $\mathbf{0.574}$ \\
\addlinespace[2pt]
\emph{Target-only oracle (reference)} & \emph{in-domain} & \emph{$0.440 \pm 0.029$} & \emph{$0.774 \pm 0.004$} & --- \\
\bottomrule
\end{tabular}
\end{threeparttable}
\end{table*}

Three patterns stand out. First, MELD achieves the highest F1 among all evaluated UDA methods in \emph{both} transfer directions. For Teyuna~$\rightarrow$~Miraflores, it reaches an F1 of $0.459$, compared with $0.397$ for the source-only model and $0.422$ for the strongest competing UDA baseline, MMD/DAN (single). For Miraflores~$\rightarrow$~Teyuna, it reaches $0.594$, compared with $0.557$ for source-only and $0.585$ for the strongest competing baseline, DANN (multi). Adding KNN geometric descriptors further increases F1 to $0.542$ and $0.606$, producing the strongest overall results in both directions.

Second, the competing UDA methods show no consistent improvement across transfer directions. The strongest competing baseline depends on the direction: MMD/DAN (single) performs best for Teyuna~$\rightarrow$~Miraflores ($0.422$), whereas DANN (multi) performs best for Miraflores~$\rightarrow$~Teyuna ($0.585$). However, each performs comparatively poorly in the opposite direction. Across all competing variants, the largest improvement over source-only in either direction is only $+0.028$; seven of the nine variants fall below source-only for Teyuna~$\rightarrow$~Miraflores and three of nine do so for Miraflores~$\rightarrow$~Teyuna. Increasing the number of adapted feature levels also has no consistent effect for these baselines: multi-level adaptation benefits DANN and Deep CORAL in some cases, but degrades MMD/DAN in both transfer directions.

Third, the transfer problem is strongly asymmetric. The source-only model reaches F1 $0.557$ when evaluated on Teyuna but only $0.397$ when evaluated on Miraflores, while the corresponding target-only oracles reach $0.774$ and $0.440$, respectively. 

\subsection{Comparison with the in-domain reference}
\label{sec:results_oracle}

The target-only oracle provides an in-domain reference for interpreting the absolute scores (Table~\ref{tab:main_results}, final row). For Miraflores~$\rightarrow$~Teyuna, supervised training directly on Teyuna reaches an F1 of $0.774$, whereas MELD and MELD + geometry attain approximately $77\%$ and $78\%$ of this reference, respectively. Thus, although the adapted models recover a substantial fraction of the performance achievable with direct target domain supervision, a gap to fully supervised in-domain training remains when sufficient target labels are available. For Teyuna~$\rightarrow$~Miraflores, however, the target-only oracle is weaker ($0.440$). This may partly reflect the much lower prevalence of terrace examples in Miraflores (\(\approx1.5\%\) areal coverage versus \(\approx6.5\%\) in Teyuna; Section~\ref{sec:data}), which provides fewer positive examples for supervised training. The asymmetry may also be related to differences in the expression of terrace morphology between the two regions. Although Miraflores and Teyuna share structural similarities, the terraces at Teyuna provide a richer and more clearly expressed representation of terrace structure, whereas those at Miraflores are generally more degraded and less morphologically distinct. Training on Teyuna may therefore expose the model to a broader set of terrace characteristics that remain informative for detecting the more weakly expressed features in Miraflores. In this direction, the MELD variants, which use no Miraflores labels during training or model selection, achieve F1 scores of \(0.459\) without geometry and \(0.542\) with geometry, both matching or exceeding the target-only in-domain reference of \(0.440\). By contrast, all competing UDA methods remain below this reference, with the strongest baseline reaching an F1 of \(0.422\). This suggests that, when target-domain annotations are scarce, transferring supervision from a more richly labelled source region while exploiting unlabeled target tiles can be competitive with limited in-domain supervision, and that among the evaluated adaptation approaches only MELD achieves this level of performance.

\subsection{Diffusion-level ablation}
\label{sec:results_levels}

Table~\ref{tab:layer_ablation} isolates the contribution of activating diffusion at different PointNet++ levels, using coordinate-only inputs. Refining only the deepest level $l=4$ gives a small gain over source-only, most clearly for Miraflores~$\rightarrow$~Teyuna. Adding level $l=3$ helps Teyuna~$\rightarrow$~Miraflores but is less stable in the reverse direction, while adding level $l=2$ recovers much of the reverse-transfer performance. In both directions the strongest result is obtained when all four levels are active. This pattern is consistent with the motivation for multi-level refinement: shallower levels carry local geometric information and deeper levels encode coarser semantic context, and terrace transfer appears to require both.

\begin{table*}[t]
\centering
\caption{MELD level ablation: effect of activating diffusion adaptation at different PointNet++ feature levels (no KNN features; adaptation enabled). The first column lists the active level set. F1 as mean $\pm$ standard deviation over three random seeds. Best result per column in bold.}
\label{tab:layer_ablation}
\begin{tabular}{lcc}
\toprule
\textbf{Active diffusion levels} & \textbf{\begin{tabular}[c]{@{}c@{}}Teyuna $\rightarrow$\\ Miraflores\end{tabular}} & \textbf{\begin{tabular}[c]{@{}c@{}}Miraflores $\rightarrow$\\ Teyuna\end{tabular}} \\
\midrule
$\{4\}$ & $0.408 \pm 0.037$ & $0.564 \pm 0.007$ \\
$\{3,4\}$ & $0.443 \pm 0.019$ & $0.534 \pm 0.025$ \\
$\{2,3,4\}$ & $0.440 \pm 0.057$ & $0.586 \pm 0.030$ \\
$\{1,2,3,4\}$ & $\mathbf{0.459 \pm 0.038}$ & $\mathbf{0.594 \pm 0.015}$ \\
\bottomrule
\end{tabular}
\end{table*}

\subsection{Precision--recall analysis}
\label{sec:pr_analysis}

\begin{figure*}[t]
    \centering
    \includegraphics[width=0.88\textwidth]{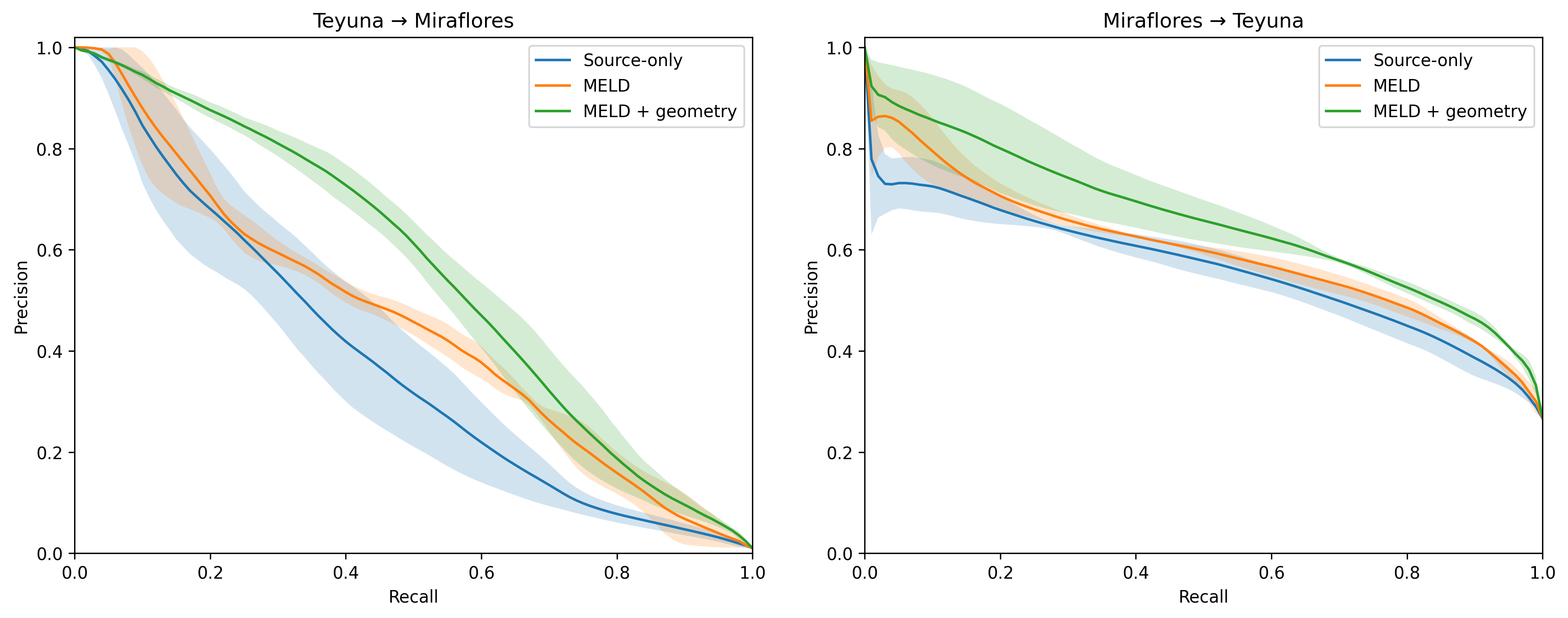}
    \caption{Precision--recall curves for the three principal configurations, averaged
over three random seeds. The left panel shows Teyuna
$\rightarrow$ Miraflores and the right panel Miraflores
$\rightarrow$ Teyuna. Solid lines show the mean across three random seeds, and shaded regions indicate one standard deviation across seeds. The compared
configurations are source-only PointNet++, MELD, and MELD+geometry.}
    \label{fig:curves}
\end{figure*}

Figure~\ref{fig:curves} compares the precision-recall behaviour of the
three principal configurations: the source-only PointNet++ model, MELD, and MELD + KNN geometry. Curves are averaged over the three
random seeds, and shaded regions indicate one standard deviation.

Precision--recall analysis is particularly informative for this task because
terrace points constitute a minority class, especially in Miraflores. For
Teyuna $\rightarrow$ Miraflores, average precision increases from
$0.381\pm0.070$ for the source-only model to $0.454\pm0.014$ with MELD and to $0.556\pm0.028$ with MELD+geometry. The
improvement is visible over a broad range of recall values rather than at a
single operating point. At recall 0.5, mean precision increases from
approximately 0.316 for source-only to 0.457 with MELD and 0.612 with MELD + geometry. Thus, for a comparable proportion of
recovered terrace points, the adapted models produce fewer false-positive
detections, with the geometry-enhanced model giving the strongest precision--recall trade-off.

The same ordering is observed for Miraflores $\rightarrow$ Teyuna. Average
precision increases from $0.563\pm0.015$ for source-only to
$0.597\pm0.011$ with MELD and $0.657\pm0.038$ with MELD+geometry. At recall 0.5, the corresponding mean precisions are
approximately 0.578, 0.598, and 0.657. The improvement due to diffusion
adaptation is therefore present in both transfer directions, while the
additional benefit of explicit local geometry is particularly pronounced in
the Teyuna $\rightarrow$ Miraflores direction.

These results are consistent with the fixed-threshold F1 comparison in
Table~\ref{tab:main_results} and show that MELD generally improves the mean precision--recall trade-off across a wide range of operating conditions, rather than only at the probability threshold of 0.5. This is operationally relevant for
archaeological prospection, where the decision threshold can be adjusted
depending on whether the objective is to recover as many possible terrace
locations as possible or to restrict expert inspection to a smaller number of
higher-confidence candidates.

\section{Discussion}
\label{sec:discussion}

\paragraph{Two complementary sources of information} The results indicate that cross-region terrace segmentation benefits from two complementary cues. First, target domain latent diffusion improves over the source-only baseline in both transfer directions, showing that unlabeled target point clouds carry distributional information that can be exploited without target labels. Second, KNN geometric descriptors add further gains when combined with adaptation: terraces are geometric landforms defined by local slope changes, planar treads, and aspect-related morphology, so explicit normal and slope cues complement the learned PointNet++ features.

\paragraph{Explicit  target distribution modelling outperforms marginal alignment} Classical UDA methods reduce source--target discrepancy through moment matching, correlation alignment, adversarial domain confusion, or target-side regularisation \citep{long2015dan,sun2016coral,sun2016deepcoral,ganin2016dann,long2018cdan,grandvalet2004}. These approaches have been successful across many vision settings and remain central to the UDA literature \citep{wang2018survey}. In the present cross-region terrace task, however, the conventional UDA baselines were less effective and markedly less consistent than MELD. For Teyuna~$\rightarrow$~Miraflores, single-level MMD/DAN was the strongest competing baseline, whereas for Miraflores~$\rightarrow$~Teyuna the strongest baseline was multi-level DANN. In both transfer directions, however, MELD achieved the highest F1 among the evaluated adaptation methods. Relative to the source-only PointNet++ model, MELD improved F1 in both transfer directions, whereas the competing approaches did not show the same consistency: several configurations fell below the source-only baseline, particularly for Teyuna~$\rightarrow$~Miraflores, and the identity of the best conventional method changed with transfer direction. The contrast is also evident when adaptation is extended across the PointNet++ hierarchy. For the alignment-based methods, moving from a single deep adaptation level to multiple levels produced mixed outcomes: multi-level DANN improved in the Miraflores~$\rightarrow$~Teyuna direction, Deep CORAL showed only modest changes, and multi-level MMD/DAN degraded in both directions. By contrast, the diffusion ablation shows an overall strengthening of performance as  target distribution refinement is extended through the hierarchy, with adaptation at all four levels yielding the best no-geometry result in both directions. Consequently, MELD obtains the highest F1 among the evaluated UDA methods for both transfer directions. We attribute this pattern to the different adaptation mechanism: rather than only reducing an aggregate discrepancy between source and target representations, the diffusion model explicitly learns a denoising prior from unlabelled target latent features and uses that prior to refine the representations supplied to the segmentation decoder. This interpretation is consistent with recent studies that have used diffusion models to address distribution shift through diffusion-driven adaptation, generative transformation, or diffusion-derived representations \citep{gao2023dda,shim2024cloudfixer,peng2024dad,ji2025diff}. Taken together, the results suggest that, for this cross-region point cloud task, relying primarily on global source--target alignment is less effective than explicitly modelling the structure of the target feature distribution. Target distribution modelling therefore provides a distinct adaptation mechanism in which the target domain is not treated only as a distribution to be aligned with, but as a distribution whose latent structure can be learned and used directly to guide feature refinement.

\paragraph{Cross-region predictions can support archaeological prospection by reducing the search space} Although MELD improves cross-region detection, the best F1 values remain below perfect segmentation, understood here as the theoretical case in which all terrace and non-terrace points are classified correctly. This is unsurprising for subtle, eroded, and vegetation-affected landforms transferred across regions without access to target labels. Previous studies of archaeological machine learning and LiDAR analysis have highlighted the value of automated methods as tools for directing expert attention toward likely archaeological features \citep{verschoofvandervaart2019,mazzacca2022seeing}. In this context, the per-point terrace probabilities can be used to rank candidate detections for visual inspection or field verification. The precision-recall behaviour in Figure~\ref{fig:curves} shows whether the model can maintain useful precision while recovering a substantial proportion of terrace points, which is important when predictions are used to prioritise candidate locations for expert inspection. This is particularly useful when obtaining target-region labels is expensive, while complete unlabelled LiDAR coverage is already available.

\paragraph{Cross-region supervision can be competitive with scarce in-domain supervision}
The comparison with the target-only reference also highlights the potential value of transferring labels across archaeological regions when annotations in the target region are scarce. For Teyuna $\rightarrow$ Miraflores, the target-only model trained directly from Miraflores labels achieves an F1 score of $0.440$, whereas MELD reaches $0.459$ without target labels and $0.542$ when geometry is included. This result should not be interpreted as evidence that unsupervised adaptation is generally superior to target domain supervision. Rather, the target-only reference is constrained by the comparatively small number of terrace examples available in Miraflores, while the adapted model can exploit labelled examples from the more terrace-rich Teyuna region together with the complete unlabelled Miraflores feature distribution. Cross-region learning can therefore provide a useful form of information sharing when the target region contains too few labelled positive examples to support a strong supervised model by itself. In archaeological applications, this may be particularly important because the regions for which automated mapping is most valuable may also be those for which only limited expert annotation has been completed.

\section{Limitations and future work}
\label{sec:limitation}
\paragraph{The current evaluation is transductive} This setting is chosen because it reflects the practical archaeological scenario in which a complete unlabelled LiDAR survey of a new region is available at once and can be exploited to adapt the model before large-scale prospection within that survey. This is appropriate when a new archaeological LiDAR survey is already available and the objective is to improve detection within that survey. However, it does not show whether the adapted model will generalise to target samples that were not seen during adaptation, which is another practical scenario where only part of a survey is used for adaptation and the model is then deployed on newly acquired or previously unseen areas. This distinction between transductive adaptation and inductive adaptation has been discussed in previous domain adaptation studies \citep{mirrashed2013domain,yan2022dual}.

\paragraph{Inductive adaptation is an important next step} Future work should evaluate the method in an inductive setting, where one unlabelled portion of the target region is used for adaptation and a spatially separate portion is reserved for testing. This would test whether the diffusion model learns a reusable target domain prior rather than benefiting mainly from access to the specific samples being predicted. Such a setting could also arise in archaeological prospection workflows, where part of a survey may be available for adaptation before the model is applied to newly processed or subsequently acquired areas.

\paragraph{Domain generalisation offers a more challenging deployment scenario} Domain generalisation removes access to the target region entirely during model development and requires direct application to an unseen domain \citep{muandet2013domain,zhou2023domain}. For archaeological site detection, this could enable a detector trained across several regions to be applied immediately to new LiDAR surveys without target-specific adaptation. This is particularly attractive for large-scale prospection, but also more difficult because domain shifts may arise from differences in vegetation, terrain, preservation, point density, and sensor characteristics. Related domain shifts caused by differences in LiDAR sensor configuration, sampling sparsity, and scene distribution have been reported for LiDAR semantic segmentation \citep{kim2023single}.

\paragraph{Future work should compare all three settings} Transductive adaptation, inductive adaptation, and domain generalisation represent different levels of dependence on target data. Transductive adaptation can specialise most strongly to a known survey, inductive adaptation tests whether this knowledge transfers to unseen samples from the same target domain, and domain generalisation aims for deployment without any prior target access. Evaluating these settings under a common protocol would clarify how much performance is traded for greater independence from target data and whether diffusion-based target modelling can be extended toward more broadly reusable archaeological detectors.

\paragraph{The study considers a single archaeological feature class}
The present study is limited to regions in which terraces are the sole archaeological feature of interest and the segmentation task is therefore formulated as a binary distinction between terrace and non-terrace points. In other archaeological landscapes, however, the features of interest may be more heterogeneous and may include multiple classes with different geometric characteristics, spatial scales, preservation states, and relationships to the surrounding terrain. Such heterogeneity could make cross-region adaptation more challenging because the target domain feature distribution would contain several distinct archaeological modes. Future work should therefore asses MELD in multi-class archaeological settings containing a broader range of feature types.

\paragraph{Richer geometric descriptors may provide further complementary gains} The present geometry-enhanced configuration uses surface normals, slope, and aspect, which primarily describe local surface orientation. The additional gains obtained from these descriptors suggest that richer explicit geometric representations may provide further complementary information. Future work could therefore investigate descriptors characterising other aspects of terrace morphology, including local curvature or surface variation, roughness, changes in surface orientation, aspect variability, and relative elevation within the surrounding neighbourhood. Such descriptors could capture surface irregularity, breaks in slope, local topographic position, and spatial consistency that are not explicitly represented by the orientation-based descriptors considered here.

\section{Conclusion}
\label{sec:conclusion}

We presented MELD, a transductive multi-level embedding-space latent-diffusion framework for unsupervised domain adaptation of point-cloud terrace detection. MELD uses PointNet++ for hierarchical segmentation, trains per-level diffusion denoisers on unlabeled target latent features, and refines features before decoding so that the representation the decoder consumes is rendered more target-like. In bidirectional transfer between two Colombian regions, MELD improves over a source-only baseline in both directions and attains the highest F1 of ten compared UDA configurations. KNN-derived local geometry adds further gains, and a level ablation indicates that refining all PointNet++ levels gives the strongest result. Relative to a target-only in-domain reference, the adapted models reach $77$--$78\%$ of the supervised in-domain performance where labels are plentiful and achieve comparable or higher performance where target-region labels are scarce. The framework therefore provides a practical route to target-aware archaeological mapping when labels exist in one region but only unlabeled point clouds are available in another.

\section*{CRediT authorship contribution statement}

\textbf{Sina Mohammadi:} Conceptualization, Data curation, Formal analysis, Investigation, Methodology, Software, Validation, Visualization, Writing original draft, Writing review \& editing.
\textbf{Roderik Lindenbergh:} Funding acquisition Resources, Writing - review \& editing.
\textbf{Javier Ruiz-Pérez}: Project administration, Writing review \& editing.
\textbf{Marco Madella:} Project administration, Supervision, Writing review \& editing. 
\textbf{Sebastian Fajardo:} Conceptualization, Data curation, Funding acquisition, Investigation, Methodology, Project administration, Software, Supervision, Validation, Writing review \& editing.



\section*{Acknowledgements}

This research is part of the "Mapping the Archaeological Pre-Columbian Heritage in South America – MAPHSA" Project funded by Arcadia – a charitable fund of Lisbet Rausing and Peter Baldwin. This research was also supported by a Leiden-Delft-Erasmus Global Support Grant (Grant No. LDE-127).

\bibliographystyle{elsarticle-harv}
\bibliography{references}

@inproceedings{qi2017pointnet,
  title={PointNet: Deep Learning on Point Sets for 3D Classification and Segmentation},
  author={Qi, Charles R. and Su, Hao and Mo, Kaichun and Guibas, Leonidas J.},
  booktitle={Proceedings of the IEEE Conference on Computer Vision and Pattern Recognition},
  pages={652--660},
  year={2017}
}

@inproceedings{qi2017pointnetplusplus,
  title={PointNet++: Deep Hierarchical Feature Learning on Point Sets in a Metric Space},
  author={Qi, Charles R. and Yi, Li and Su, Hao and Guibas, Leonidas J.},
  booktitle={Advances in Neural Information Processing Systems},
  volume={30},
  year={2017}
}

@inproceedings{ho2020ddpm,
  title={Denoising Diffusion Probabilistic Models},
  author={Ho, Jonathan and Jain, Ajay and Abbeel, Pieter},
  booktitle={Advances in Neural Information Processing Systems},
  volume={33},
  pages={6840--6851},
  year={2020}
}

@inproceedings{long2015dan,
  title={Learning Transferable Features with Deep Adaptation Networks},
  author={Long, Mingsheng and Cao, Yue and Wang, Jianmin and Jordan, Michael I.},
  booktitle={Proceedings of the 32nd International Conference on Machine Learning},
  pages={97--105},
  year={2015}
}

@article{ganin2016dann,
  title={Domain-Adversarial Training of Neural Networks},
  author={Ganin, Yaroslav and Ustinova, Evgeniya and Ajakan, Hana and Germain, Pascal and Larochelle, Hugo and Laviolette, Fran\c{c}ois and Marchand, Mario and Lempitsky, Victor},
  journal={Journal of Machine Learning Research},
  volume={17},
  number={59},
  pages={1--35},
  year={2016}
}

@inproceedings{long2018cdan,
  title={Conditional Adversarial Domain Adaptation},
  author={Long, Mingsheng and Cao, Zhangjie and Wang, Jianmin and Jordan, Michael I.},
  booktitle={Advances in Neural Information Processing Systems},
  volume={31},
  pages={1640--1650},
  year={2018}
}

@article{stular2021airborne,
  title={Airborne LiDAR Point Cloud Processing for Archaeology. Pipeline and QGIS Toolbox},
  author={\v{S}tular, Benjamin and Eichert, Stefan and Lozi\'{c}, Edisa},
  journal={Remote Sensing},
  volume={13},
  number={16},
  pages={3225},
  year={2021},
  doi={10.3390/rs13163225}
}

@article{osorio2023beyond,
  title={Beyond Visualization: Remote Sensing Applications in Prehispanic Settlements to Understand Ancient Anthropogenic Land Use and Occupation in the Sierra Nevada de Santa Marta, Colombia},
  author={Rodr\'{i}guez Osorio, Daniel and Giraldo, Santiago and Mazuera, Eduardo and Burbano, Andr\'{e}s and Figueredo, Estefan\'{i}a},
  journal={Latin American Antiquity},
  volume={34},
  number={4},
  pages={821--841},
  year={2023},
  doi={10.1017/laq.2022.91}
}

@article{mazzacca2022seeing,
  title={Seeing Among Foliage with LiDAR and Machine Learning: Towards a Transferable Archaeological Pipeline},
  author={Mazzacca, G. and Grilli, E. and Cirigliano, G. P. and Remondino, F. and Campana, S.},
  journal={The International Archives of the Photogrammetry, Remote Sensing and Spatial Information Sciences},
  volume={XLVI-2/W1-2022},
  pages={365--372},
  year={2022},
  doi={10.5194/isprs-archives-XLVI-2-W1-2022-365-2022}
}

@article{yang2023heritagepointcloud,
  title={Three-Dimensional Point Cloud Semantic Segmentation for Cultural Heritage: A Comprehensive Review},
  author={Yang, Su and Hou, Miaole and Li, Songnian},
  journal={Remote Sensing},
  volume={15},
  number={3},
  pages={548},
  year={2023},
  doi={10.3390/rs15030548}
}

@article{tiwari2023waterharvesting,
  title={A Deep Learning Approach for Automatic Identification of Ancient Agricultural Water Harvesting Systems},
  author={Tiwari, Arti and Silver, Micha and Karnieli, Arnon},
  journal={International Journal of Applied Earth Observation and Geoinformation},
  volume={118},
  pages={103270},
  year={2023},
  doi={10.1016/j.jag.2023.103270}
}

@article{tubog2025terraces,
  title={Detection of Agricultural Terraces Platforms Using Machine Learning from Orthophotos and LiDAR-Based Digital Terrain Model: A Case Study in Roya Valley of Southeast France},
  author={Tubog, Michael Vincent and Emsellem, Karine and Bouissou, Stephane},
  journal={Land},
  volume={14},
  number={5},
  pages={962},
  year={2025},
  doi={10.3390/land14050962}
}

@inproceedings{grandvalet2004,
  title={Semi-Supervised Learning by Entropy Minimization},
  author={Grandvalet, Yves and Bengio, Yoshua},
  booktitle={Advances in Neural Information Processing Systems},
  volume={17},
  pages={529--536},
  year={2004}
}

@inproceedings{french2018selfensembling,
  title={Self-Ensembling for Visual Domain Adaptation},
  author={French, Geoffrey and Mackiewicz, Michal and Fisher, Mark},
  booktitle={International Conference on Learning Representations},
  year={2018}
}

@inproceedings{zou2018cbst,
  title={Unsupervised Domain Adaptation for Semantic Segmentation via Class-Balanced Self-Training},
  author={Zou, Yang and Yu, Zhiding and Kumar, B. V. K. Vijaya and Wang, Jinsong},
  booktitle={European Conference on Computer Vision},
  pages={289--305},
  year={2018}
}

@inproceedings{qin2019pointdan,
  title={{PointDAN}: A Multi-Scale {3D} Domain Adaption Network for Point Cloud Representation},
  author={Qin, Can and You, Haoxuan and Wang, Lichen and Kuo, C.-C. Jay and Fu, Yun},
  booktitle={Advances in Neural Information Processing Systems},
  volume={32},
  pages={7192--7203},
  year={2019}
}

@inproceedings{sohldickstein2015,
  title={Deep Unsupervised Learning using Nonequilibrium Thermodynamics},
  author={Sohl-Dickstein, Jascha and Weiss, Eric A. and Maheswaranathan, Niru and Ganguli, Surya},
  booktitle={Proceedings of the 32nd International Conference on Machine Learning},
  pages={2256--2265},
  year={2015}
}

@inproceedings{sun2016coral,
  title={Return of Frustratingly Easy Domain Adaptation},
  author={Sun, Baochen and Feng, Jiashi and Saenko, Kate},
  booktitle={Proceedings of the AAAI Conference on Artificial Intelligence},
  volume={30},
  number={1},
  year={2016}
}

@inproceedings{sun2016deepcoral,
  title={Deep {CORAL}: Correlation Alignment for Deep Domain Adaptation},
  author={Sun, Baochen and Saenko, Kate},
  booktitle={European Conference on Computer Vision Workshops},
  pages={443--450},
  year={2016}
}

@article{verschoofvandervaart2019,
  title={Learning to Look at {LiDAR}: The Use of {R-CNN} in the Automated Detection of Archaeological Objects in {LiDAR} Data from the Netherlands},
  author={Verschoof-van der Vaart, Wouter B. and Lambers, Karsten},
  journal={Journal of Computer Applications in Archaeology},
  volume={2},
  number={1},
  pages={31--40},
  year={2019},
  doi={10.5334/jcaa.32}
}

@article{wang2018survey,
  title={Deep Visual Domain Adaptation: A Survey},
  author={Wang, Mei and Deng, Weihong},
  journal={Neurocomputing},
  volume={312},
  pages={135--153},
  year={2018}
}

@inproceedings{gao2023dda,
  title={Back to the Source: Diffusion-Driven Adaptation to Test-Time Corruption},
  author={Gao, Jin and Zhang, Jialing and Liu, Xihui and Darrell, Trevor and Shelhamer, Evan and Wang, Dequan},
  booktitle={Proceedings of the IEEE/CVF Conference on Computer Vision and Pattern Recognition},
  pages={11786--11796},
  year={2023}
}

@inproceedings{shim2024cloudfixer,
  title={CloudFixer: Test-Time Adaptation for 3D Point Clouds via Diffusion-Guided Geometric Transformation},
  author={Shim, Hajin and Kim, Changhun and Yang, Eunho},
  booktitle={European Conference on Computer Vision},
  pages={454--471},
  year={2024}
}

@article{peng2024dad,
  title={Unsupervised Domain Adaptation via Domain-Adaptive Diffusion},
  author={Peng, Duo and Ke, Qiuhong and Ambikapathi, ArulMurugan and Yazici, Yasin and Lei, Yinjie and Liu, Jun},
  journal={IEEE Transactions on Image Processing},
  volume={33},
  pages={4245--4260},
  year={2024},
  doi={10.1109/TIP.2024.3424985}
}

@inproceedings{ji2025diff,
  title={Diffusion Features to Bridge Domain Gap for Semantic Segmentation},
  author={Ji, Yuxiang and He, Boyong and Qu, Chenyuan and Tan, Zhuoyue and Qin, Chuan and Wu, Liaoni},
  booktitle={IEEE International Conference on Acoustics, Speech and Signal Processing (ICASSP)},
  pages={1--5},
  year={2025},
  doi={10.1109/ICASSP49660.2025.10888537}
}

@inproceedings{mirrashed2013domain,
  title     = {Domain Adaptive Classification},
  author    = {Mirrashed, Fatemeh and Rastegari, Mohammad},
  booktitle = {Proceedings of the IEEE International Conference on Computer Vision},
  pages     = {2608--2615},
  year      = {2013},
  doi       = {10.1109/ICCV.2013.324}
}

@inproceedings{yan2022dual,
  title     = {Dual Moving Average Pseudo-Labeling for Source-Free Inductive Domain Adaptation},
  author    = {Yan, Hao and Guo, Yuhong},
  booktitle = {British Machine Vision Conference},
  year      = {2022}
}

@inproceedings{muandet2013domain,
  title     = {Domain Generalization via Invariant Feature Representation},
  author    = {Muandet, Krikamol and Balduzzi, David and Sch{\"o}lkopf, Bernhard},
  booktitle = {Proceedings of the 30th International Conference on Machine Learning},
  volume    = {28},
  pages     = {10--18},
  year      = {2013},
  publisher = {PMLR}
}

@article{zhou2023domain,
  title   = {Domain Generalization: A Survey},
  author  = {Zhou, Kaiyang and Liu, Ziwei and Qiao, Yu and Xiang, Tao and Loy, Chen Change},
  journal = {IEEE Transactions on Pattern Analysis and Machine Intelligence},
  volume  = {45},
  number  = {4},
  pages   = {4396--4415},
  year    = {2023},
  doi     = {10.1109/TPAMI.2022.3195549}
}

@inproceedings{kim2023single,
  title     = {Single Domain Generalization for LiDAR Semantic Segmentation},
  author    = {Kim, Hyeonseong and Kang, Yoonsu and Oh, Changgyoon and Yoon, Kuk-Jin},
  booktitle = {Proceedings of the IEEE/CVF Conference on Computer Vision and Pattern Recognition},
  pages     = {17587--17598},
  year      = {2023},
  doi       = {10.1109/CVPR52729.2023.01687}
}

@article{freeland2016automated,
  title={Automated feature extraction for prospection and analysis of monumental earthworks from aerial LiDAR in the Kingdom of Tonga},
  author={Freeland, Travis and Heung, Brandon and Burley, David V and Clark, Geoffrey and Knudby, Anders},
  journal={Journal of Archaeological Science},
  volume={69},
  pages={64--74},
  year={2016},
  publisher={Elsevier}
}

@article{trier2019using,
  title={Using deep neural networks on airborne laser scanning data: Results from a case study of semi-automatic mapping of archaeological topography on Arran, Scotland},
  author={Trier, {\O}ivind Due and Cowley, David C and Waldeland, Anders Ueland},
  journal={Archaeological Prospection},
  volume={26},
  number={2},
  pages={165--175},
  year={2019},
  publisher={Wiley Online Library}
}

@article{verschoof2020combining,
  title={Combining deep learning and location-based ranking for large-scale archaeological prospection of LiDAR data from the Netherlands},
  author={Verschoof-van der Vaart, Wouter B and Lambers, Karsten and Kowalczyk, Wojtek and Bourgeois, Quentin PJ},
  journal={ISPRS International Journal of Geo-Information},
  volume={9},
  number={5},
  pages={293},
  year={2020},
  publisher={MDPI}
}

@article{https://doi.org/10.1002/arp.1806,
author = {Bonhage, Alexander and Eltaher, Mahmoud and Raab, Thomas and Breuß, Michael and Raab, Alexandra and Schneider, Anna},
title = {A modified Mask region-based convolutional neural network approach for the automated detection of archaeological sites on high-resolution light detection and ranging-derived digital elevation models in the North German Lowland},
journal = {Archaeological Prospection},
volume = {28},
number = {2},
pages = {177-186},
doi = {https://doi.org/10.1002/arp.1806},
url = {https://onlinelibrary.wiley.com/doi/abs/10.1002/arp.1806},
eprint = {https://onlinelibrary.wiley.com/doi/pdf/10.1002/arp.1806},
year = {2021}
}

@inproceedings{vu2019advent,
  title={Advent: Adversarial entropy minimization for domain adaptation in semantic segmentation},
  author={Vu, Tuan-Hung and Jain, Himalaya and Bucher, Maxime and Cord, Matthieu and P{\'e}rez, Patrick},
  booktitle={2019 IEEE/CVF conference on computer vision and pattern recognition (CVPR)},
  pages={2512--2521},
  year={2019},
  organization={IEEE}
}

@article{prabhudesai2023diffusion,
  title={Diffusion-tta: Test-time adaptation of discriminative models via generative feedback},
  author={Prabhudesai, Mihir and Ke, Tsung-Wei and Li, Alex and Pathak, Deepak and Fragkiadaki, Katerina},
  journal={Advances in Neural Information Processing Systems},
  volume={36},
  pages={17567--17583},
  year={2023}
}

@article{liu2021adversarial,
  title={Adversarial unsupervised domain adaptation for 3D semantic segmentation with multi-modal learning},
  author={Liu, Wei and Luo, Zhiming and Cai, Yuanzheng and Yu, Ying and Ke, Yang and Junior, Jos{\'e} Marcato and Gon{\c{c}}alves, Wesley Nunes and Li, Jonathan},
  journal={ISPRS Journal of Photogrammetry and Remote Sensing},
  volume={176},
  pages={211--221},
  year={2021},
  publisher={Elsevier}
}

@article{wang2025test,
  title={Test-time adaptation for geospatial point cloud semantic segmentation with distinct domain shifts},
  author={Wang, Puzuo and Yao, Wei and Shao, Jie and He, Zhiyi},
  journal={ISPRS Journal of Photogrammetry and Remote Sensing},
  volume={229},
  pages={422--435},
  year={2025},
  publisher={Elsevier}
}

@article{li2020dance,
  title={DANCE-NET: Density-aware convolution networks with context encoding for airborne LiDAR point cloud classification},
  author={Li, Xiang and Wang, Lingjing and Wang, Mingyang and Wen, Congcong and Fang, Yi},
  journal={ISPRS Journal of Photogrammetry and Remote Sensing},
  volume={166},
  pages={128--139},
  year={2020},
  publisher={Elsevier}
}

@article{lin2020active,
  title={Active and incremental learning for semantic ALS point cloud segmentation},
  author={Lin, Yaping and Vosselman, George and Cao, Yanpeng and Yang, Michael Ying},
  journal={ISPRS journal of photogrammetry and remote sensing},
  volume={169},
  pages={73--92},
  year={2020},
  publisher={Elsevier}
}

@article{vinci2025,
author = {Vinci, Giacomo and Vanzani, Federica and Fontana, Alessandro and Campana, Stefano},
title = {LiDAR Applications in Archaeology: A Systematic Review},
journal = {Archaeological Prospection},
volume = {32},
number = {1},
pages = {81-101},
doi = {10.1002/arp.1931},
url = {https://onlinelibrary.wiley.com/doi/abs/10.1002/arp.1931},
eprint = {https://onlinelibrary.wiley.com/doi/pdf/10.1002/arp.1931},
year = {2025}
}

@article{haznedar2023implementing,
  title={Implementing PointNet for point cloud segmentation in the heritage context},
  author={Haznedar, Bulent and Bayraktar, Rabia and Ozturk, Ali Emre and Arayici, Yusuf},
  journal={Heritage Science},
  volume={11},
  number={1},
  pages={2},
  year={2023},
  publisher={Nature Publishing Group}
}

@article{pierdicca2020point,
  title={Point cloud semantic segmentation using a deep learning framework for cultural heritage},
  author={Pierdicca, Roberto and Paolanti, Marina and Matrone, Francesca and Martini, Massimo and Morbidoni, Christian and Malinverni, Eva Savina and Frontoni, Emanuele and Lingua, Andrea Maria},
  journal={Remote Sensing},
  volume={12},
  number={6},
  pages={1005},
  year={2020},
  publisher={MDPI}
}

\end{document}